\documentclass[times,twocolumn,final]{elsarticle}

\usepackage{framed,multirow}

\usepackage{amssymb}
\usepackage{amsmath}
\usepackage{graphicx}
\usepackage{array}
\usepackage{pbox}
\usepackage{algorithm}
\usepackage[noend]{algpseudocode}

\algnewcommand\algorithmicforeach{\textbf{for each}}
\algdef{S}[FOR]{ForEach}[1]{\algorithmicforeach\ #1\ \algorithmicdo}

\newcommand{\sphere}{\mathcal{S}^{2}}
\newcommand{\snm}{}

\journal{Computer Methods and Programs in Biomedicine}

\begin{document}


\begin{frontmatter}

\title{Reproducible and Interpretable Spiculation Quantification for Lung Cancer Screening}

\author[1,2]{Wookjin \snm{Choi}\fnref{fn1}}
\author[1]{Saad \snm{Nadeem}\fnref{fn1}\corref{cor1}}
\author[1]{Sadegh R. \snm{Alam}}
\author[1]{Joseph O. \snm{Deasy}}
\author[3]{Allen \snm{Tannenbaum}}
\author[1]{Wei \snm{Lu}} 
\fntext[fn1]{\textbf{The first two authors contributed equally to this work.}}
\cortext[cor1]{Corresponding author: nadeems@mskcc.org}


\address[1]{Department of Medical Physics, Memorial Sloan Kettering Cancer Center, 1275 York Ave, New York, NY 10065, USA}
\address[2]{Department of Engineering and Computer Science, Virginia State University, 1 Hayden St, Petersburg, VA 23806, USA}
\address[3]{Departments of Computer Science and Applied Mathematics \& Statistics, Stony Brook University, Stony Brook, NY 11790, USA}


\begin{abstract}
Spiculations are important predictors of lung cancer malignancy, which are spikes on the surface of the pulmonary nodules. In this study, we proposed an interpretable and parameter-free technique to quantify the spiculation using area distortion metric obtained by the conformal (angle-preserving) spherical parameterization. We exploit the insight that for an angle-preserved spherical mapping of a given nodule, the corresponding negative area distortion precisely characterizes the spiculations on that nodule. We introduced novel spiculation scores based on the area distortion metric and spiculation measures. We also semi-automatically segment lung nodule (for reproducibility) as well as vessel and wall attachment to differentiate the real spiculations from lobulation and attachment. A simple pathological malignancy prediction model is also introduced. We used the publicly-available LIDC-IDRI dataset pathologists (strong-label) and radiologists (weak-label) ratings to train and test radiomics models containing this feature, and then externally validate the models. We achieved AUC$=$0.80 and 0.76, respectively, with the models trained on the 811 weakly-labeled LIDC datasets and tested on the 72 strongly-labeled LIDC and 73 LUNGx datasets; the previous best model for LUNGx had AUC$=$0.68. The number-of-spiculations feature was found to be highly correlated (Spearman's rank correlation coefficient $\rho = 0.44$) with the radiologists' spiculation score. We developed a reproducible and interpretable, parameter-free technique for quantifying spiculations on nodules. The spiculation quantification measures was then applied to the radiomics framework for pathological malignancy prediction with reproducible semi-automatic segmentation of nodule. Using our interpretable features (size, attachment, spiculation, lobulation), we were able to achieve higher performance than previous models. In the future, we will exhaustively test our model for lung cancer screening in the clinic.
\end{abstract}


\end{frontmatter}

\section{Introduction}
Lung cancer is the most common cause of cancer-related death in the United States \citep{siegel2016cancer}. Lung cancer screening with a low-dose computed tomography (CT) for current and former smokers has been shown a clear survival benefit by the National Lung Cancer Screening Trial \citep{aberle2013nejm}. Recently radiomics studies have been proposed for various clinical applications \citep{buty2016characterization,choi2018medphy,hawkins2016predicting}, which extract a vast number of quantitative image features and then perform data mining to predict tumor responses and patient outcomes for more reliable and accurate prediction of local control and overall survival. Refer to \cite{thawani2018radiomics} for an exhaustive review of radiomics and radiogenomics studies to predict clinical outcomes in lung cancer.

The radiomics analysis has also been studied for lung cancer screening. Hawkins et al.~\cite{hawkins2016predicting} proposed a random forest classifier using 23 stable radiomic features. Buty et al.~\cite{buty2016characterization} introduced a random forest classifier using a pre-trained deep neural network feature extractor (4096 appearance features), and a spherical harmonics feature extractor (400 shape features).  The spherical harmonics are a decomposition of the frequency-space basis for representing functions defined over the sphere and applicable to describe the overall shape of the object. However, it cannot provide local features for a given region on a shape (e.g., spiculation). Kumar et al.~\cite{kumar2015discovery} proposed a deep neural network model, which used 5000 features. Liu et al.~\cite{liu2016radiological} introduced a linear classifier based on 24 image traits visually scored by physicians. Choi et al.~\cite{choi2018medphy} proposed a model for predicting malignancy in pulmonary nodules using a support vector machine classifier coupled with a least absolute shrinkage and selection operator (SVM-LASSO), which only use two CT radiomic features (size and texture). While these radiomics studies have improved the accuracy of the predictions, the lack of clinical/biological interpretation of the features remains limited.

\begin{figure*}[t!]
\begin{center}
\setlength{\tabcolsep}{1pt}
\begin{tabular}{ccccc}
\includegraphics[width=0.3\textwidth]{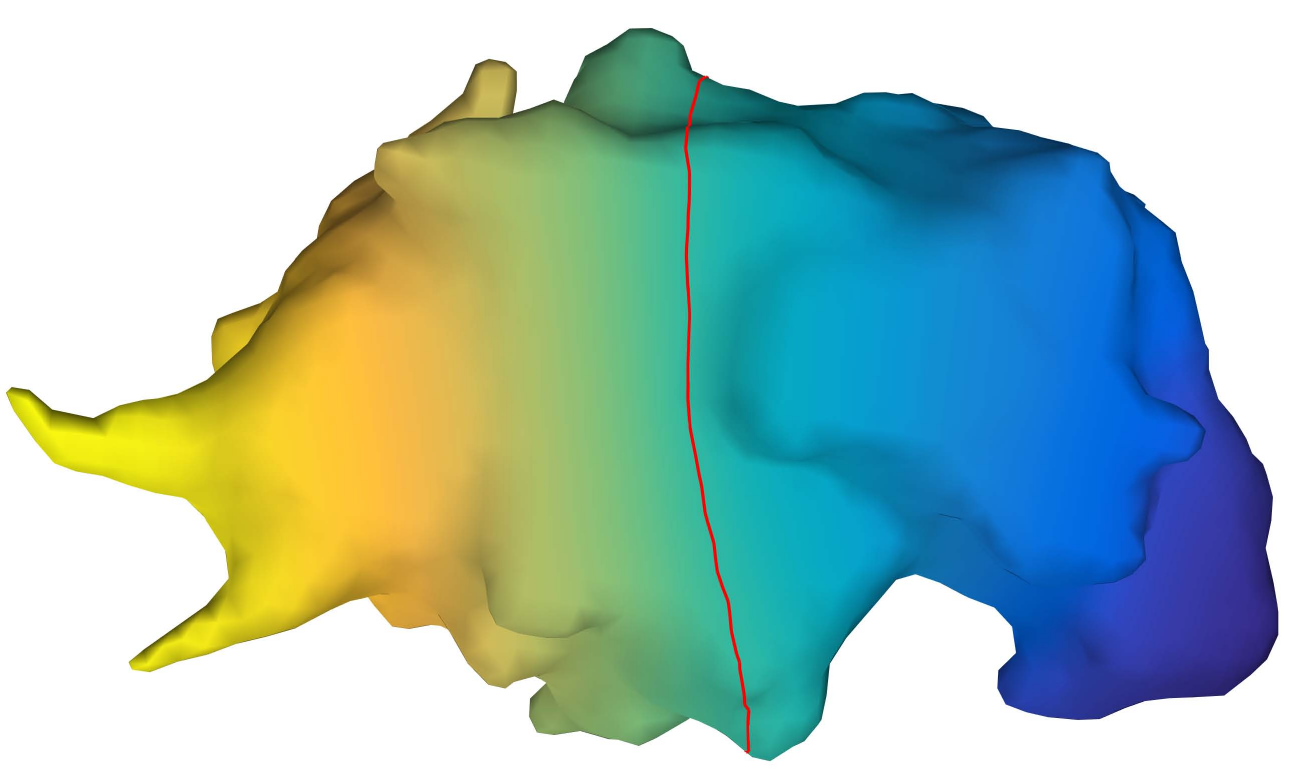}&
\includegraphics[width=0.18\textwidth]{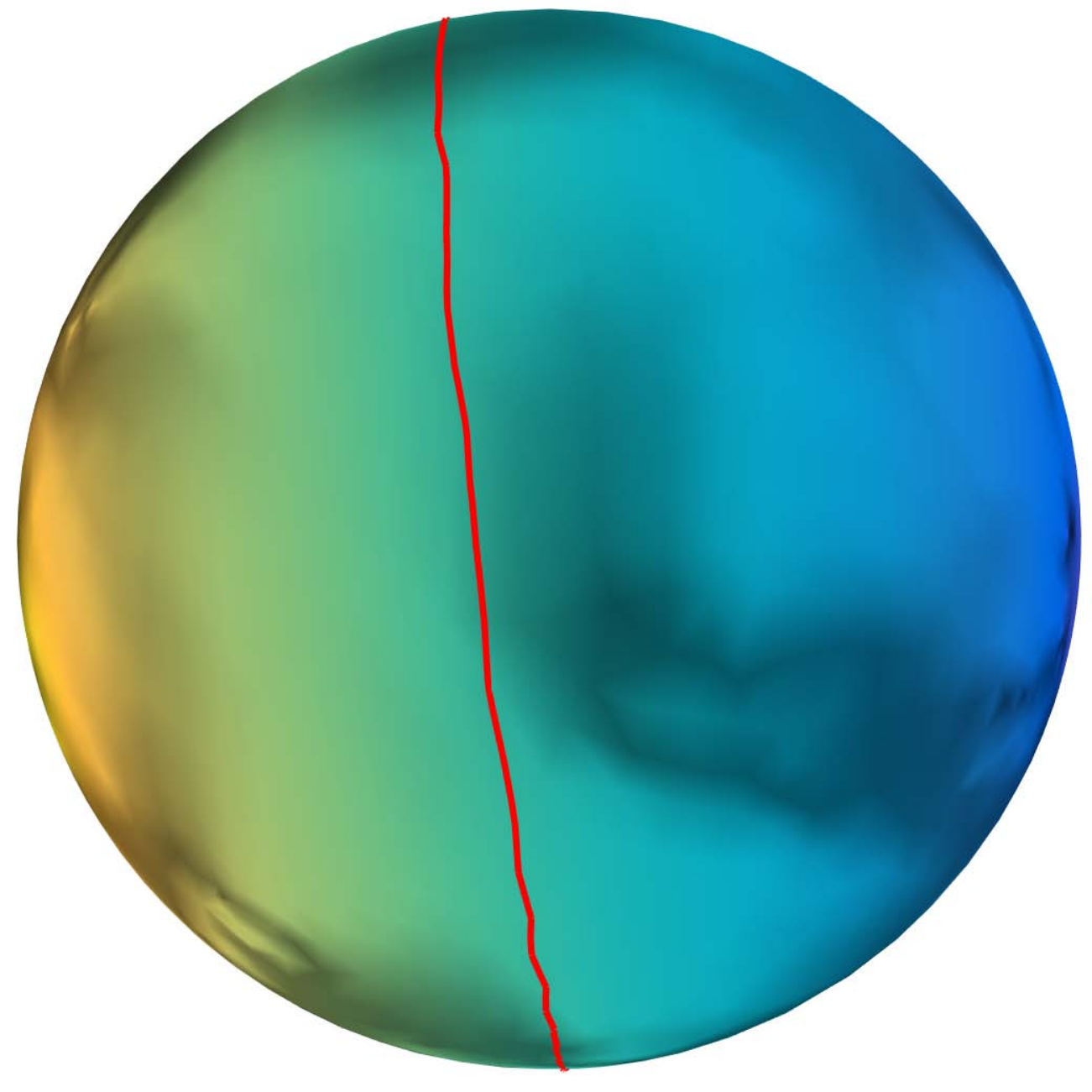}&
\includegraphics[width=0.036\textwidth]{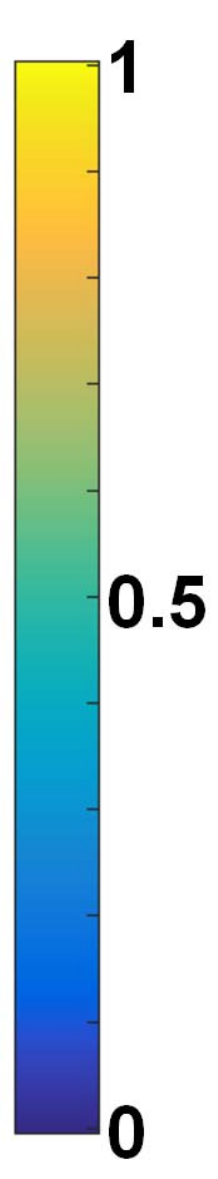}&
\includegraphics[width=0.3\textwidth]{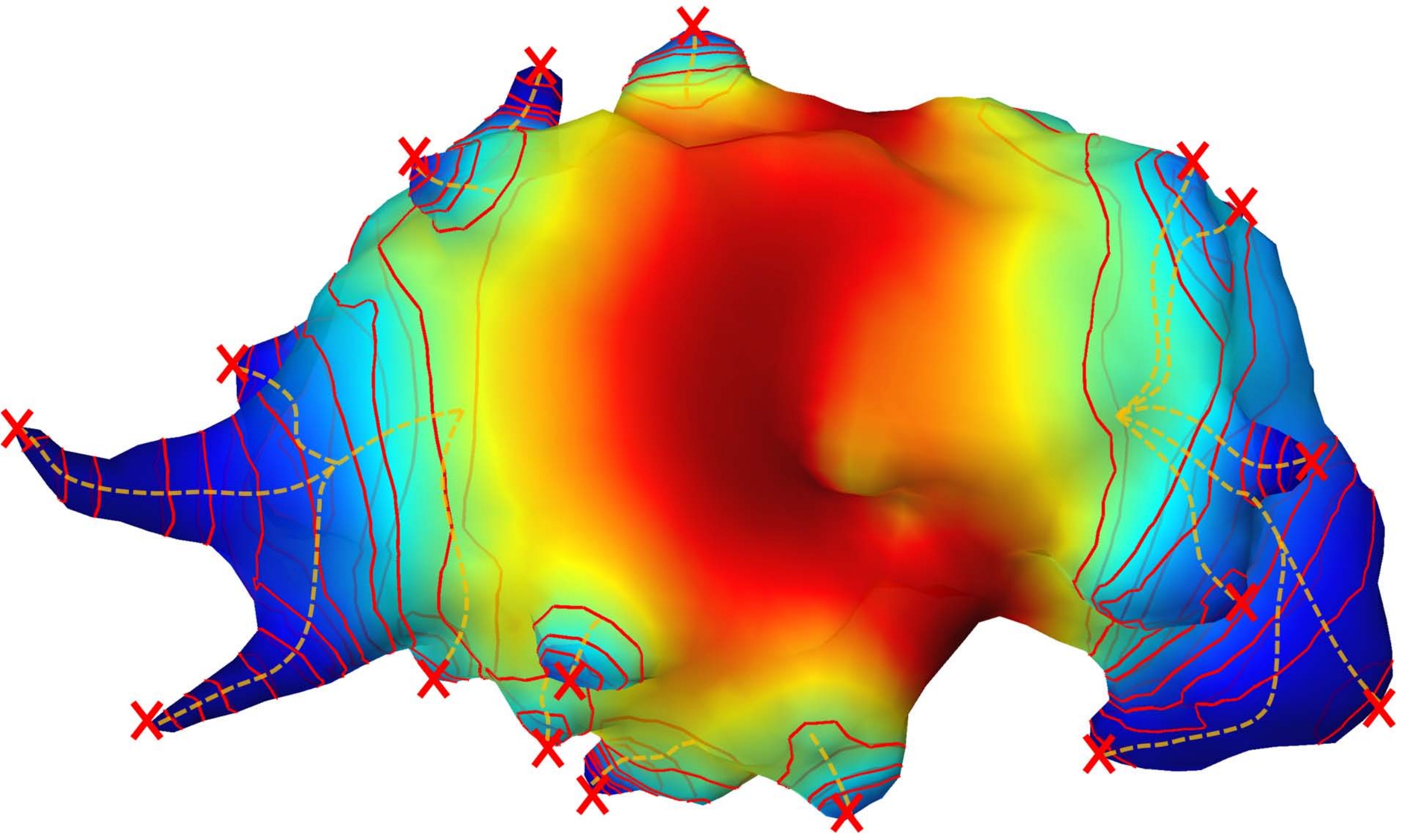}&
\includegraphics[width=0.031\textwidth]{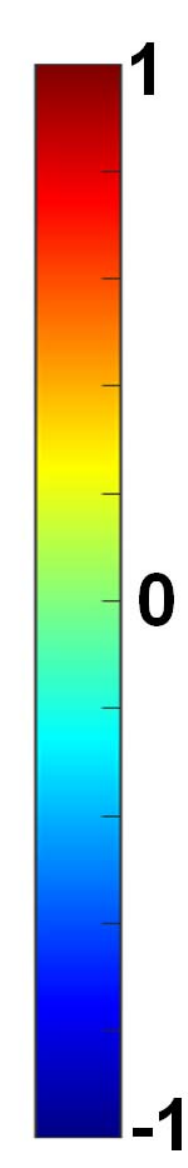}\\
(a) & (b) & \small{Eigenfunction} & (c) & \small{Area Distortion} \\
\end{tabular}
\end{center}
\caption{Spiculation quantification pipeline. (a) The first non-trivial eigenfunction of the Laplace-Beltrami operator for a given mesh is computed. The mesh was divide into two topological disks by the zeroth-level set (red curve) of this eigenfunction. The disks are conformally welded and stereographically projected to a sphere (b), in angle-preserving spherical parameterization \citep{nadeem2017spherical}. (c) The area distortion metric is applied to detect apex (red x's, which have local maximum negative area distortion), and obtain heights (yellow curves) for each spike/spiculation.
\label{fig:sphere_param}}
\end{figure*}

\begin{figure}[t!]
\begin{center}
\includegraphics[width=0.43\textwidth]{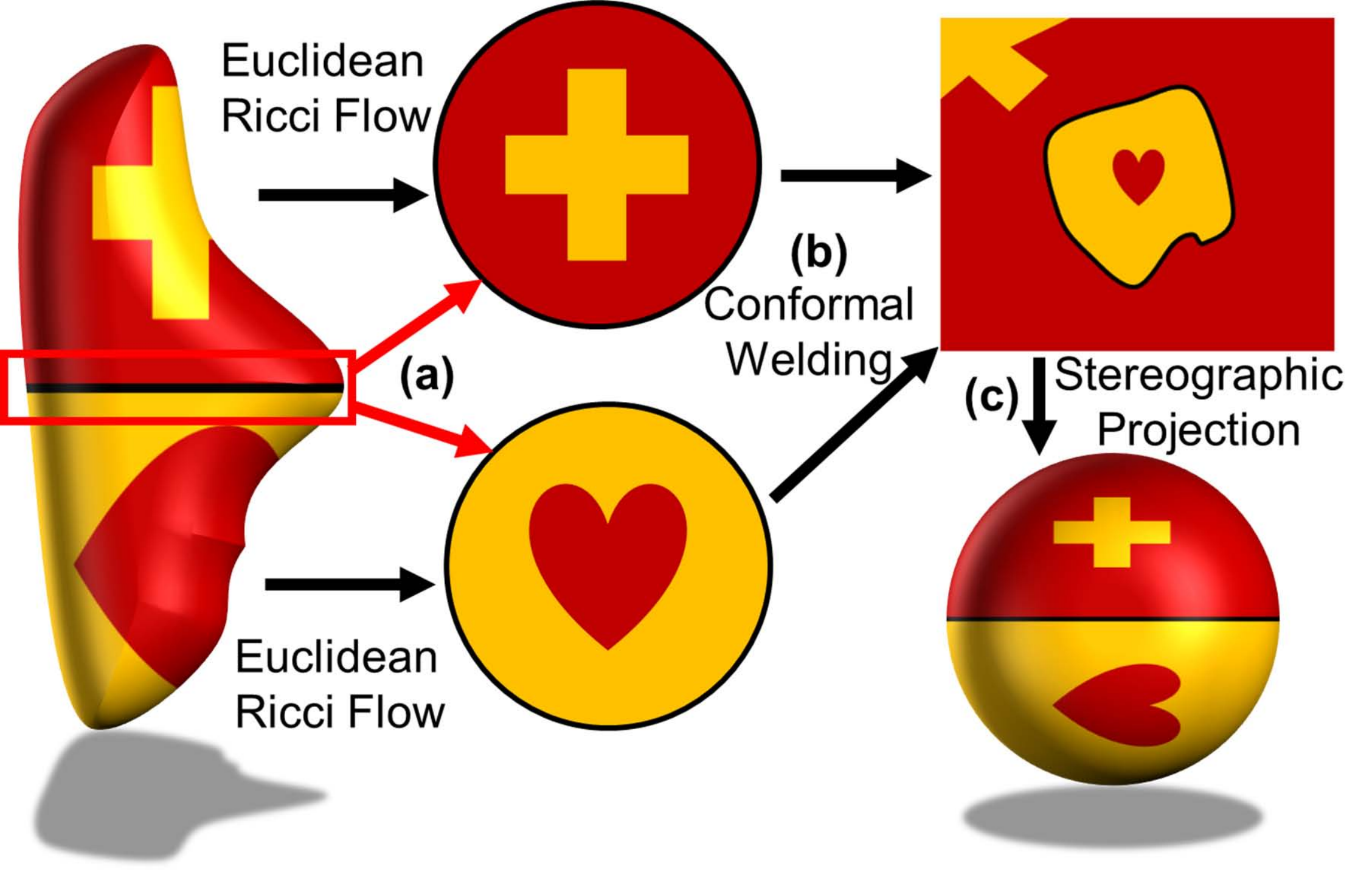}\\
\end{center}
\caption{Angle-preserving spherical parameterization schema. A given genus-0 surface is divided automatically into two topological disks via zeroth levelset of the Fiedler vector (as depicted in Figure \ref{fig:sphere_param}a). These disks are then conformally (angle-preserving) flattened via Euclidean Ricci Flow. Finally, the two flattened disks are conformally welded in an extended complex plane along their boundaries and stereographically projected to a sphere.
\label{fig:spherical_schema}}
\end{figure}

Radiographic edge characteristics of a nodule, specifically spiculation (spikes on the surface of nodules), influence the probability of malignancy \citep{swensen1997aim}. Typically, malignant nodules have blurred and irregular boundaries, while benign nodules have well-defined and smooth boundaries. The American College of Radiology (ACR) developed the Lung Imaging Reporting and Data System (Lung-RADS) to standardize the lung cancer screening on CT images using size, appearance type (spiculation, lobulation, vessel/wall attachment) and calcification \citep{mckee2015jacr}. Lung-RADS suggests spiculation as an additional image finding that increases the suspicion of malignancy to improve prediction accuracy. The McWilliams \citep{mcwilliams2013nejm} introduced a model to compute the probability of lung cancer, which uses nine variables, such as age, sex, emphysema, family history of lung cancer, the number of nodules, nodule size, nodule type, nodule location, and spiculation. Nodule size and spiculation were the significant malignancy predictors in both models.

\label{intro_spic}Spiculation quantification of pulmonary nodule has been previously studied but not in the prediction of malignancy. Niehaus et al.~\cite{niehaus2015toward} introduced a computer-aided diagnosis (CAD) system, which used the size dependence of shape features to quantify spiculations. Ciompi et al.~\cite{ciompi2015automatic} proposed a frequency-based shape descriptor specifically tailored to assess presence of spiculation in detected solid nodules for lung cancer screening. Dhara et al.~\cite{dhara2016differential} quantified spiculation peaks on a surface mesh, extracted from the binary mask of the segmented nodule. They used mean curvature and geodesic distance transformation for detecting apex of spiculation, and the baseline was then detected by tracing the sudden change of surface. The method was highly sensitive to the local variation of the surface, and hence, was challenging to detect baseline for noisy spiculation peak accurately.

In this work, we present a comprehensive pipeline to quantify spiculations, lobulations, and vessel/wall attachments, and evaluate their importance in malignancy prediction. This work extends our ShapeMI workshop paper \cite{choi2018interpretable}. The contributions of this paper are as follows:
\begin{enumerate}
     \item A novel interpretable spiculation feature is presented, computed using the area distortion metric from conformal (angle-preserving) spherical parameterization. To the best of our knowledge, we are the first ones to exploit the insight that for an angle-preserved (conformal) spherical mapping of a given nodule (e.g., using a Ricci flow algorithm \cite{nadeem2017spherical}), the corresponding negative area distortion accurately characterizes the spiculations/spikes on that nodule. Moreover, a simple one-step feature and prediction model is introduced, which only uses our interpretable features (size, spiculation, lobulation, vessel/wall attachment) and has the added advantage of using weak-labeled training data.
    \item A semi-automatic segmentation algorithm is also introduced for more accurate and reproducible lung nodule segmentation as well as vessel/wall attachment segmentation. The segmentation method leads to more accurate spiculation quantification because the attachments can be excluded from spikes on the lung nodule surface (triangular mesh) data.
    Using just our interpretable features (size, attachment, spiculation, lobulation), we were able to achieve AUC$=$0.82 on LIDC and AUC$=$0.76 on LUNGx (the previous LUNGx best being AUC$=$0.68).
    \item State-of-the-art correlation is achieved between our spiculation score (the number of spiculations $N_s$) and radiologist's spiculation score ($\rho = 0.44$).
\end{enumerate}

The paper is organized as follows: first, we introduce the spherical parameterization technique for spiculation quantification and scoring based on semi-automatic segmentation of lung nodule surface (triangular mesh) data (Sections \ref{sec:area_angle}--\ref{sec:segmentation}). The new spiculation measures are then performed for pathological malignancy prediction (Section \ref{sec:malignancy}) followed by comprehensive validation of our spiculation quantification on phantom FDA datasets from which we identify the new solid angle threshold ($T_{\Omega}$) to differentiate lobulation and spiculation in real datasets (Section \ref{sec:results:data}). The correlations between our spiculation measures and radiologist's spiculation scores (RS) are then provided along with the performance of malignancy prediction using the spiculation measures (Sections \ref{sec:results:spiculation} and \ref{sec:results:malignancy}). Finally, we discuss the limitations of our work (Section \ref{sec:discussions}) followed by conclusion (Section \ref{sec:conclusion}).

\begin{table}[b!]
\centering
\footnotesize
\caption{Symbols and their definitions}
\label{table:symbols}
\setlength{\tabcolsep}{6pt}
\begin{tabular}{lm{0.38\textwidth}}
\hline
Symbol & Definition \\
\hline
BB\_AP& Bounding box length of anterior-posterior direction\\
SD\_IDM& Standard deviation of inverse difference moment\\
$\epsilon$& Area distortion metric\\
$N_p$ & The number of peaks \\
$N_s$ & The number of spiculations \\
$N_l$ & The number of lobulations \\
$N_a$ & The number of attached peaks \\
$s_1$ \& $s_2$ & The proposed spiculation scores (sharpness and irregularity) \\
$s_a$ \& $s_b$ & Dhara's spiculation scores \cite{dhara2016differential}\\
$T_s$ & Threshold to binarize RS\\
$T_h$ & Minimum height of spiculaiton\\
$T_{\Omega}$ & Maximum solid angle of spiculation\\
sr & Steradian, the SI unit of solid angle\\
\hline
\end{tabular}
\end{table}

\section{Method}
When mapping any given compact surface (e.g., nodule) to a sphere, there is a trade-off between angle distortion and area distortion (e.g., lowering the angle distortion during the mapping increases the corresponding area distortion). \textit{Given this trade-off, the following insight can be exploited for accurately quantifying spiculations on a given nodule. For an angle-preserved (conformal) spherical mapping of a nodule (e.g., \cite{haker,nadeem2017spherical}), the negative area distortion precisely characterizes the spiculations/spikes on that nodule} (Fig. \ref{fig:sphere_param}).

\subsection{Conformal mappings and area distortion} \label{sec:area_angle}
First, we provide a theoretical overview of the area distortion in conformally mapping a genus zero Riemannian surface $S$ to the unit sphere $\sphere$ to motivate the spiculation quantification pipeline (see \cite{DoCarmo} and \cite{Tu} for the relevant mathematical background). By the {\em Theorema Egregium} of Gauss, one cannot find a diffeomorphism from $S$ with non-constant Gaussian curvature to $\sphere$ which preserves both area and angles. Furthermore, by a general result in complex analysis (uniformization), $S$ and $\sphere$ are conformally equivalent. That is, there exists a diffeomorphism $\phi: S \rightarrow \sphere$ that preserves angles. Then $\phi$ is unique up to M\"{o}bius transformation on $\sphere$. 
This is the \textbf{\emph{spherical parameterization}} of a compact genus 0 surface for which we want to measure area distortion.

One may directly use the mapping $\phi$ to compute the area distortion as in \cite{haker}. For working on a triangular mesh we have chosen the approach based on \cite{nadeem2017spherical}.
Let $g_0$ be the Riemannian metric on $S$
with corresponding Gaussian curvature $K_0$. Let $K_u$ be the curvature on the conformally equivalent surface with metric $g_u = e^{2u} g_0$. Then it is well-known (\cite{kazdan})
that
\begin{equation} \label{eq:conformal} \Delta u + K_u e^{2u} = K_0.
\end{equation}
This equation provides a specific measure of the are distortion in any spherical parameterization procedure. \cite{nadeem2017spherical} proposed a dynamic version of Eq.~\ref{eq:conformal} which is essentially the 2D Ricci Flow. Indeed, for the unit sphere $K_u=1$, and thus $u$ satisfies the Poisson equation
\begin{equation} \label{poisson} \Delta u = K_0-e^{2u}. \end{equation} $u$ is called the \emph{conformal distortion factor}, and $e^{2u}$ measures the area distortion between the surface $S$ and the sphere $\sphere$. If one examines the latter Poisson equation, one qualitatively sees that the more $K_0(x)$ varies, the greater the variation in $u$, and from the maximum principle, \textbf{\emph{spikes/spiculations may be identified by the greatest negative variation in area distortion}}.

\subsection{Angle-preserving spherical parameterization} \label{sec:conformal}

We will outline in this section the methodology proposed in \cite{nadeem2017spherical} for conformally mapping a compact genus 0 surface to a sphere, which we will use for spiculation detection/quantification.

Let $S=(V, E, F)$, denote a triangular mesh where $V$ denotes the vertices, $E$ the edges, and $F$ the faces. We assume that $S$ represents the triangulation of a genus 0 compact surface, i.e., a topological sphere. As shown in Fig. \ref{fig:spherical_schema}, the idea is to divide $S$ into two topological discs $S_1$ and $S_2$ with boundary curve given by $\gamma$. $S_1, S_2,$, and  $\gamma$ may be found via the zeroth level set of the eigenfunction corresponding to the smallest positive eigenvalue of the (discrete) Laplacian (the so-called {\em Fiedler vector}). A discretization of the 2D Ricci (Yamabe flow) is then used to conformally flatten $S_1$ and $S_2$ to discs, which are then conformally welded together and stereographically projected to get a conformal mapping to the Riemann sphere. Specific details can be found in \cite{nadeem2017spherical}.

\subsection{Spiculation detection and quantification pipeline} \label{sec:spiculation}
We now formulate the pipeline derived from overall program discussed in Sections~\ref{sec:area_angle}-\ref{sec:conformal} in a discrete setting with respect to a triangulated surface $S=(V,E,F)$. Here, one may measure the area distortion on each triangle. The spiculation quantification pipeline using this discrete version of spherical parameterization is as follows (with height and width detection; see Fig.~\ref{fig:sphere_param} and Alg. \ref{algo:peak_detection}):

\begin{algorithm}[t!]
\caption{Peak detection algorithm}\label{algo:peak_detection}
\small
\begin{algorithmic}[1]
\Function{RecursePeakContours}{$S,\epsilon,T$} \label{algo_ln:recursive}
    \State $T_n \gets T$
    \ForEach{node $t \in T$}
        \State Initialize $(S_t,\epsilon_t)$ subset of $S$ and $\epsilon$ when $\epsilon<min(\epsilon(t.b))$
        \State $B \gets$ \Call{FindBoundaries}{$S_t,\epsilon_t$} \Comment{Next level boundaries}
        \If{$B$ is empty}
            \State $T_n \gets\{T \cup node(nil,t)\}$ \Comment{Terminal node (apex)} 
        \Else
            \State $T_t \gets\{T_t \cup node(b,t)$\} for each $b \in B$ 
            \State $T_n \gets\{T \cup$ \Call{RecursePeakContours}{$S_t,\epsilon_t,T_t$}\}
        \EndIf
    \EndFor
    \Return $T_n$ \label{algo_ln:recursive_end}
\EndFunction

\Function{PeakDetection}{$S,\epsilon$}
    \State Initialize $(S_b,\epsilon_b)$ subset of $S$ and $\epsilon$ when $\epsilon<0$ \label{algo_ln:baseline}
    \State $B \gets$ \Call{FindBoundaries}{$S_b,\epsilon_b$} \Comment{Baseline detection}
    \State $T \gets\{T \cup node(b,nil)$\} for each $b \in B$ \Comment{Generate root nodes} 
    \State $T \gets$ \Call{RecursePeakContours}{$S_b,\epsilon_b,T$}
\Ensure a peak $p$ is a stack consisting of nodes from each individual terminal node (apex) to a root node (baseline) of the tree $T$.
\EndFunction

\end{algorithmic}
\end{algorithm}

\begin{enumerate}
    \item Compute conformal (angle-preserving) spherical parameterization \cite{nadeem2017spherical}: The first non-trivial eigenfunction of the Laplace-Beltrami operator is computed for a given mesh (Fig. \ref{fig:sphere_param}a). The mesh was divided into two topological disks by the zeroth-level set (red curve in Fig. \ref{fig:sphere_param}a) of this eigenfunction. The disks are conformally welded and stereographically projected to a sphere (Fig. \ref{fig:sphere_param}b).
    \item Compute the normalized area distortion. For each vertex $v_i$, the \emph{area distortion} is defined as
\[
    \epsilon_i := \log \frac{\sum_{j,k} A([\phi(v_i),\phi(v_j),\phi(v_k)])}{\sum_{j,k} A([v_i,v_j,v_k])}
\]
where $[v_i,v_j,v_k]$ is the triangle formed by ${v_i,v_j,v_k}$ and $A(.)$ represents the area of a triangle.
    \item Find all the baselines $B$ where area distortion is zero (Line \ref{algo_ln:baseline} in Algorithm~\ref{algo:peak_detection}).
    \item Recursively search closed curves from the baseline (zero area distortion) to the apex (the smallest area distortion) using the level-set method. During the search, the closed curves can break into multiple closed curves and move towards different apexes. Each pair of the apex (a terminal node) and its corresponding closed curves define a peak and are assigned unique IDs to track their progression and for height and width computations in the next step. (Line \ref{algo_ln:recursive}--\ref{algo_ln:recursive_end} in Algorithm~\ref{algo:peak_detection})
    \item Compute the sum of the distances between the successive centroids of the closed curves to obtain the peak height. The peak width is also computed on the area distortion map of the peak using a full width half maximum concept.
\end{enumerate}

\label{lobulation} Malignant nodules are more likely to have irregular, lobulated or spiculated margins due to the spread of malignant cells within the pulmonary interstitium. The peak detection is able to capture both lobules and spicules. Among them the spiculated nodule is more likely malignant than others. The classification of the detected peaks into the spiculation (sharp peak) and lobulation (curved peak) will provide more descriptive feature information for the malignancy prediction. To exclude lobulation from spiculation, we applied thresholding for the height ($T_h\ge3 mm$) and solid angle ($T_\Omega\le0.65 sr$). The solid angle threshold was suggested in \cite{dhara2016differential}, and we confirmed it with the phantom FDA dataset (see Results section). We also applied a full width half maximum concept for more robust width measures of a peak, using the peak surface and its area distortion. The peak width was measured on an iso-contour at half minimum area distortions (all negative values).

\begin{figure*}[t!]
\begin{center}
\setlength{\tabcolsep}{2pt}
\begin{tabular}{c>{\centering\arraybackslash}m{0.16\textwidth}>{\centering\arraybackslash}m{0.16\textwidth}>{\centering\arraybackslash}m{0.16\textwidth}>{\centering\arraybackslash}m{0.16\textwidth}>{\centering\arraybackslash}m{0.16\textwidth}}
(a)&
\includegraphics[width=0.16\textwidth]{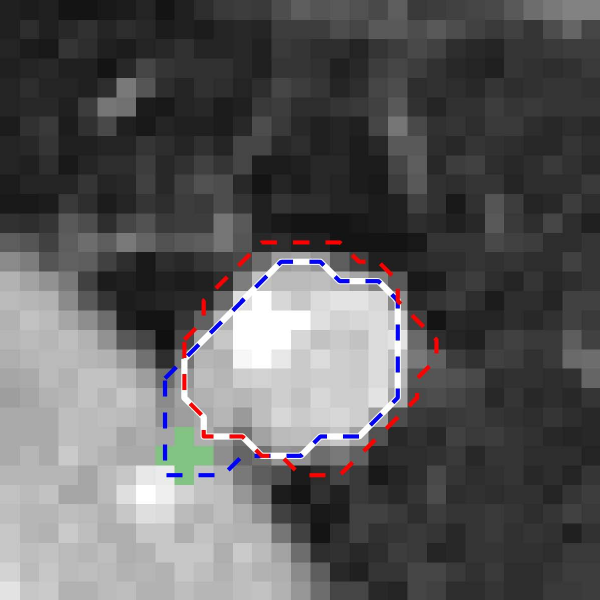}&
\includegraphics[width=0.16\textwidth]{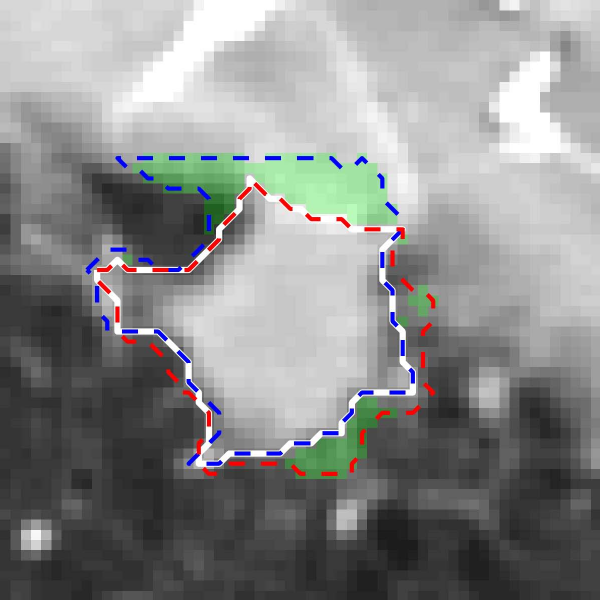}&
\includegraphics[width=0.16\textwidth]{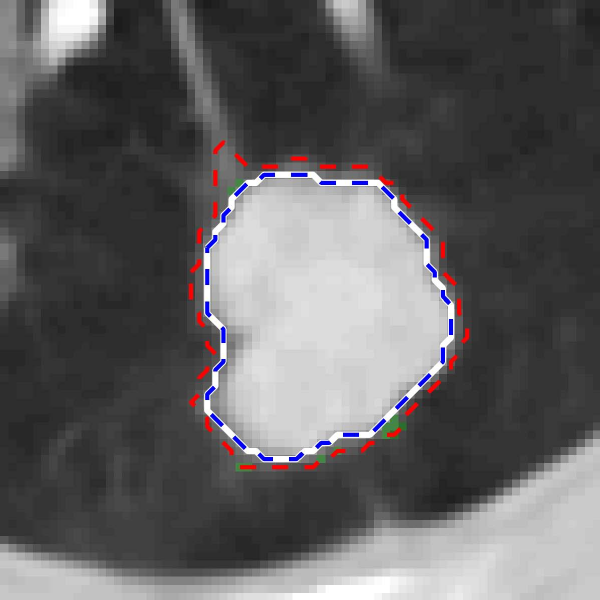}&
\includegraphics[width=0.16\textwidth]{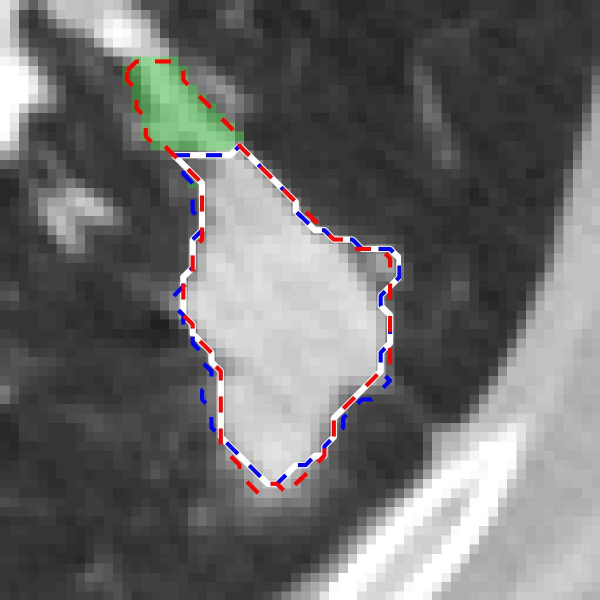}&
\includegraphics[width=0.16\textwidth]{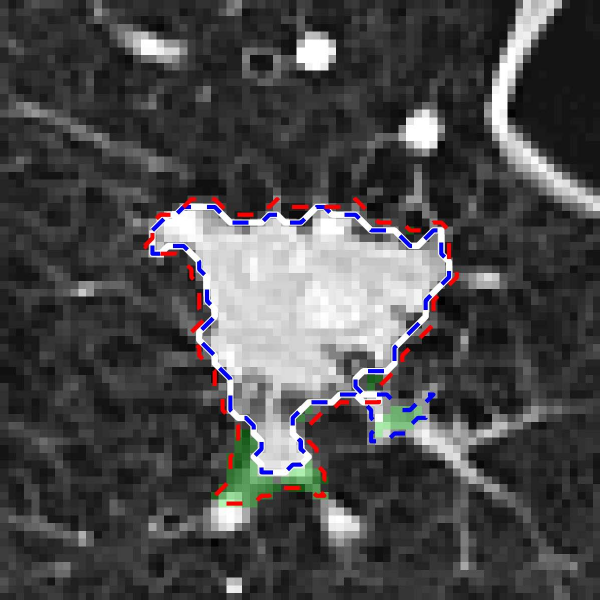}\\
(b)&
\includegraphics[width=0.16\textwidth]{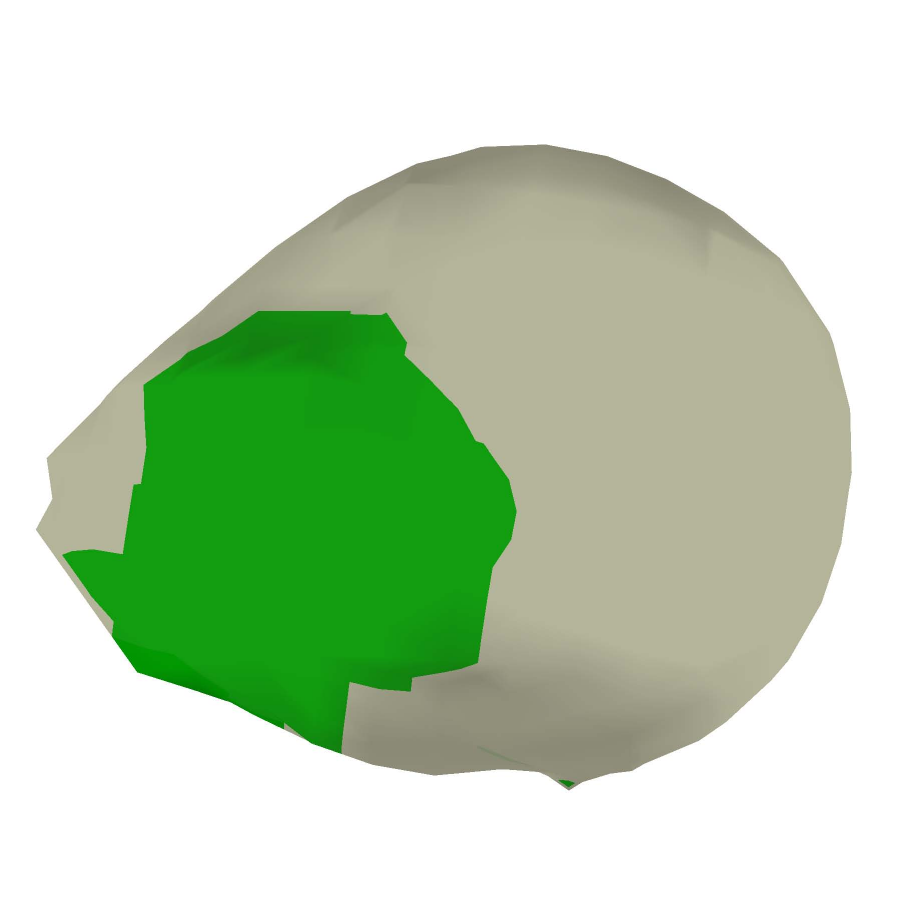}&
\includegraphics[width=0.16\textwidth]{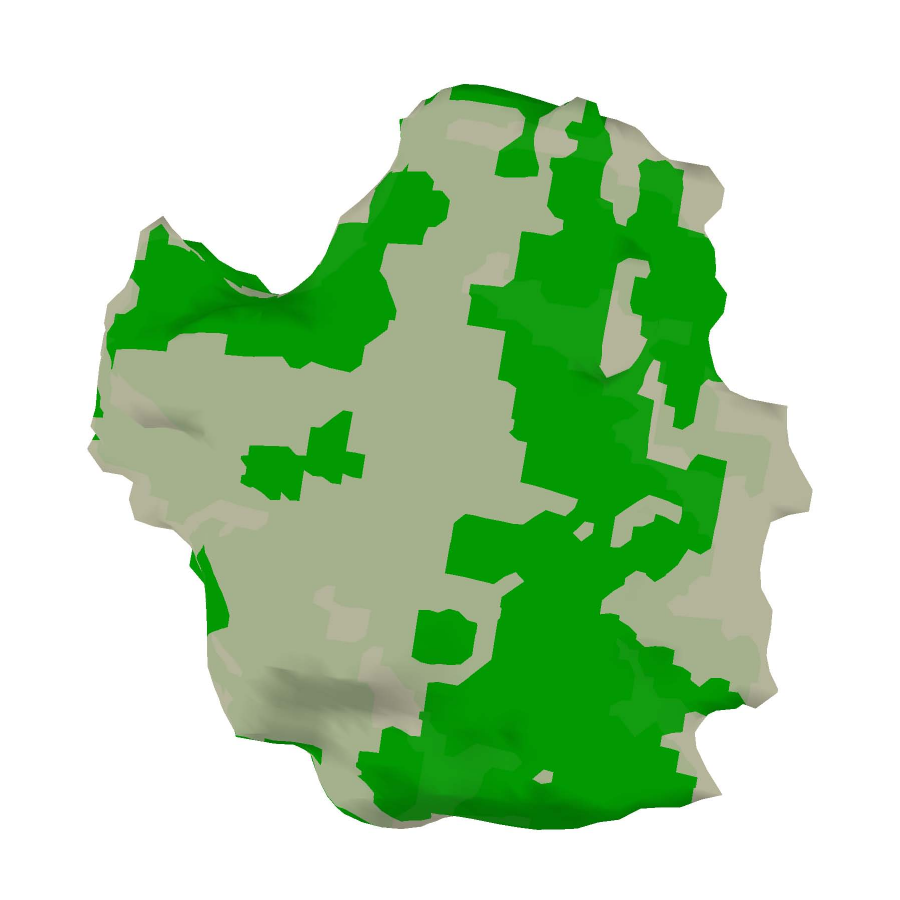}&
\includegraphics[width=0.16\textwidth]{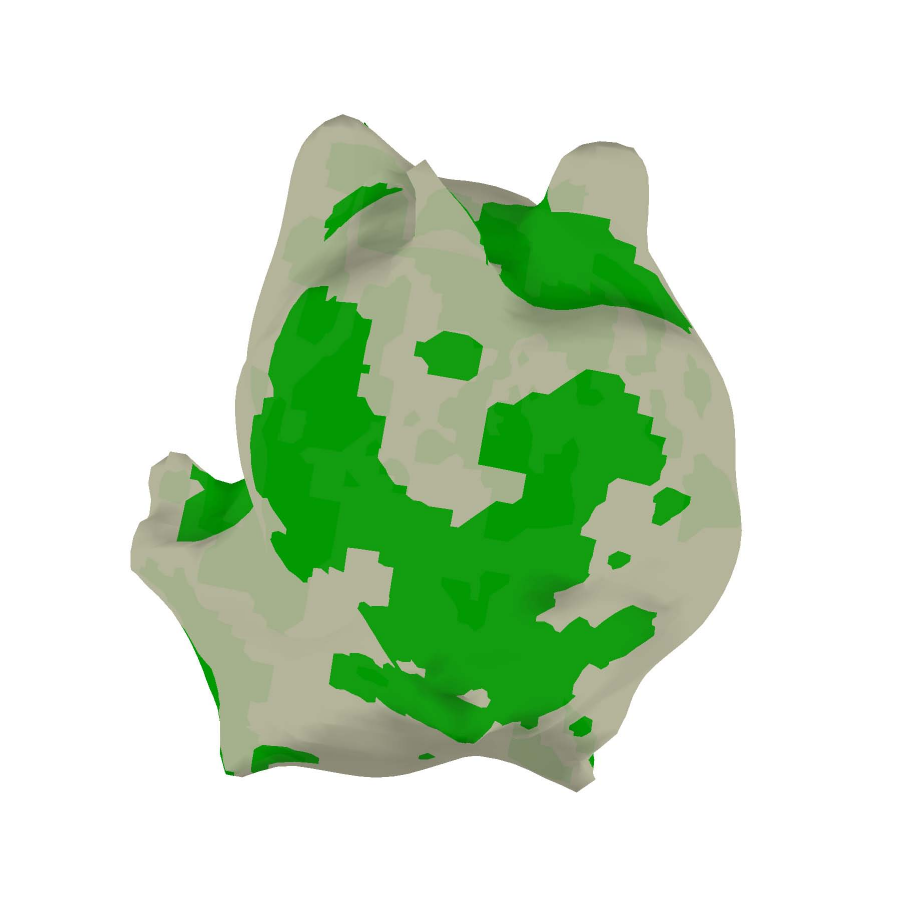}&
\includegraphics[width=0.16\textwidth]{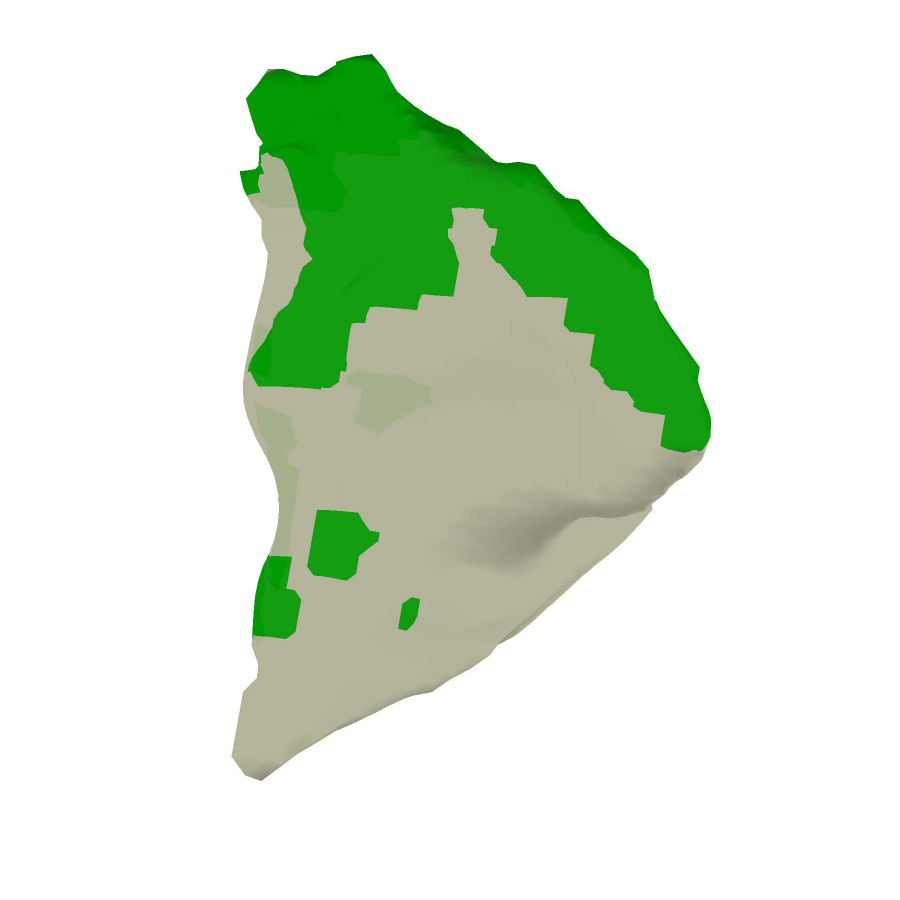}&
\includegraphics[width=0.16\textwidth]{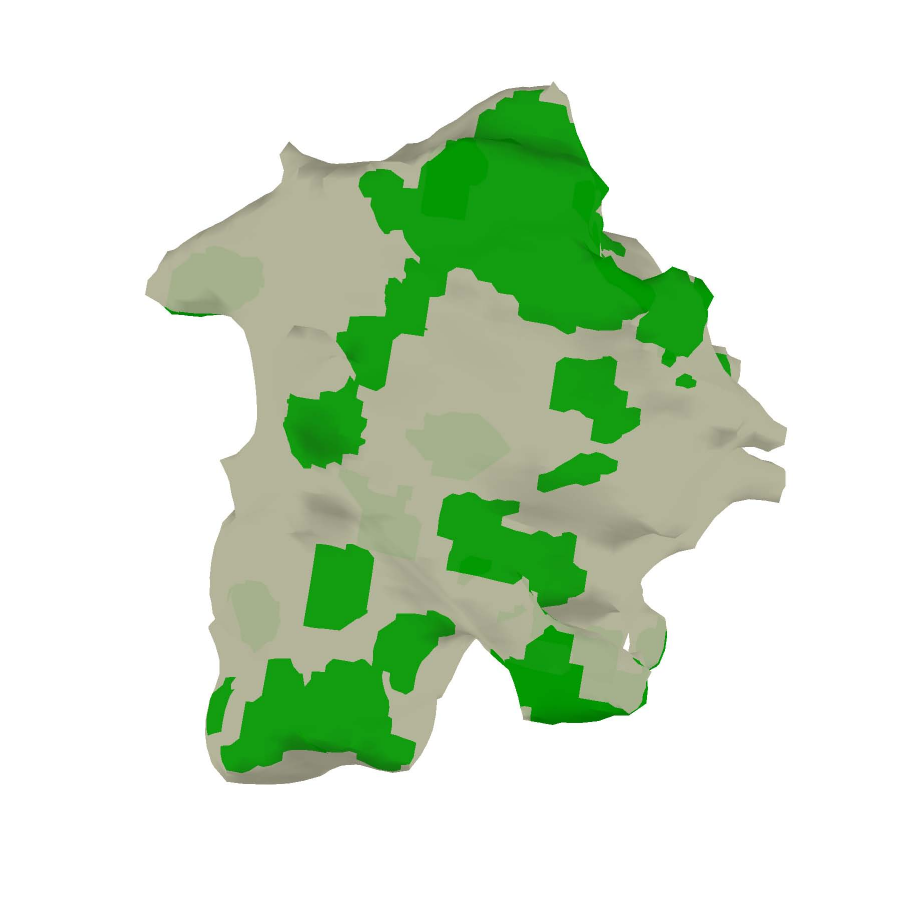}\\
(c)&
\includegraphics[width=0.16\textwidth]{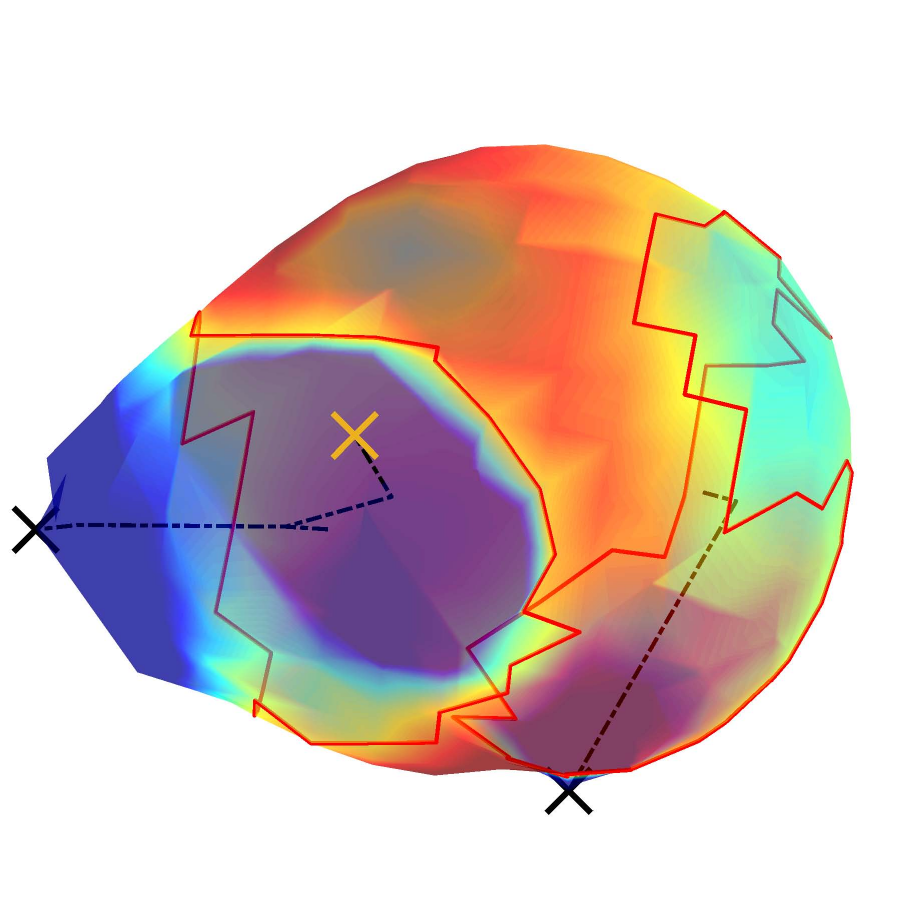}&
\includegraphics[width=0.16\textwidth]{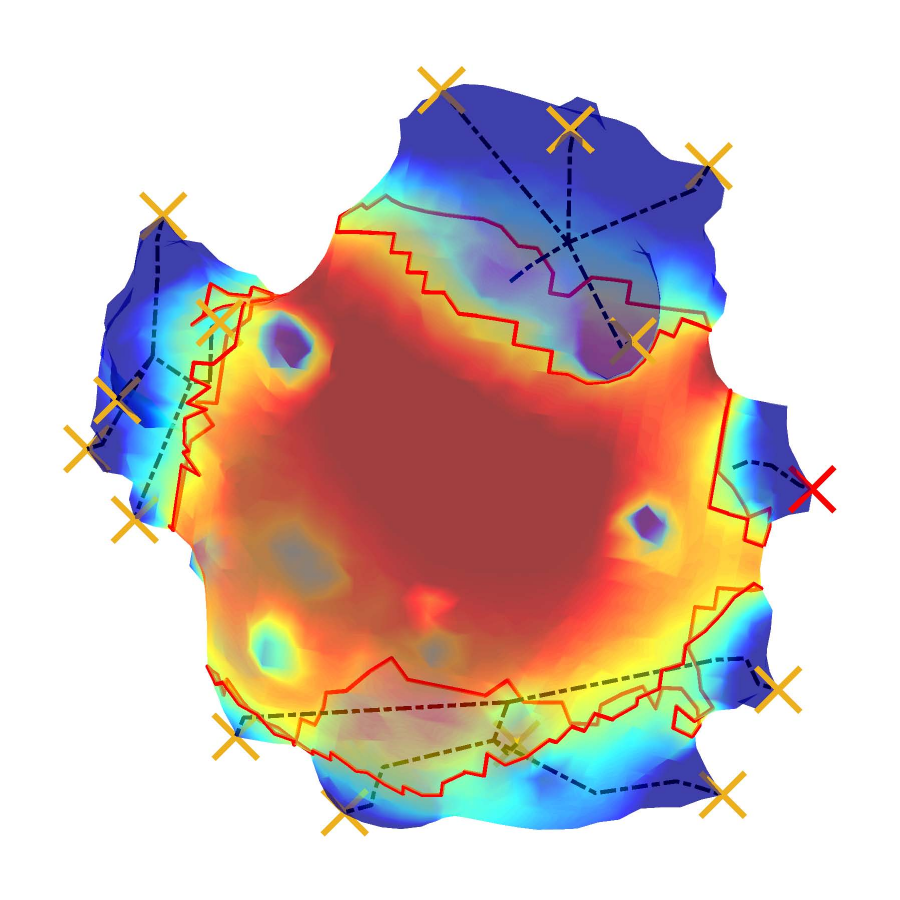}&
\includegraphics[width=0.16\textwidth]{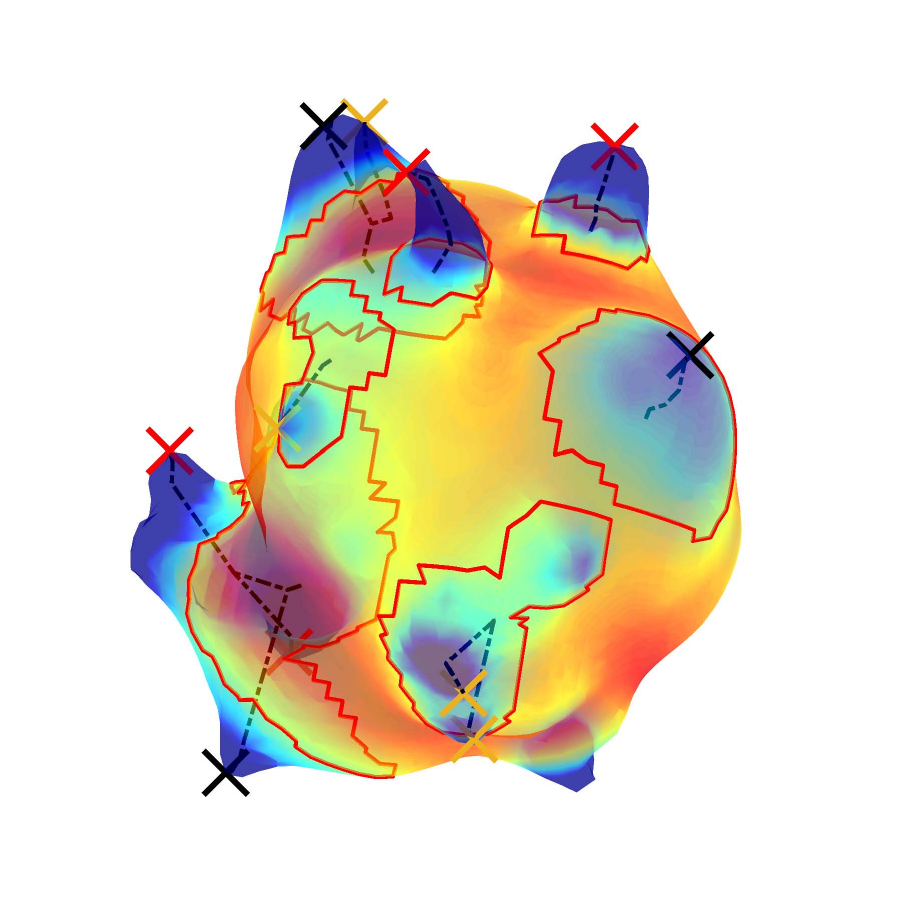}&
\includegraphics[width=0.16\textwidth]{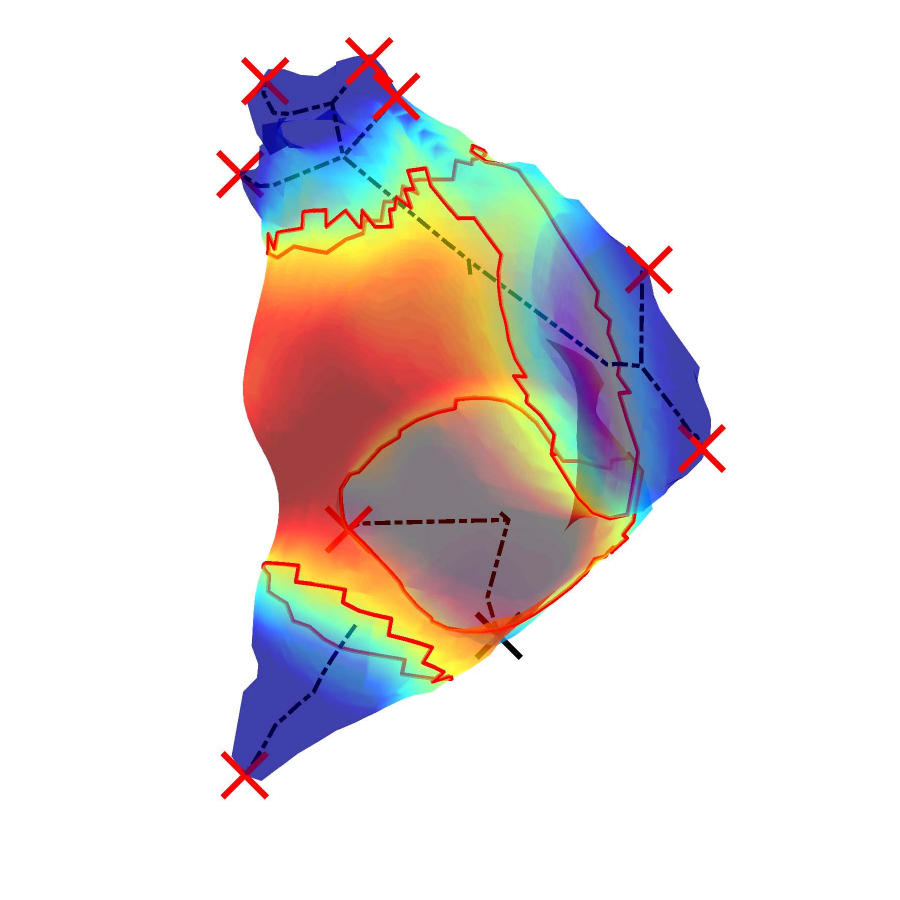}&
\includegraphics[width=0.16\textwidth]{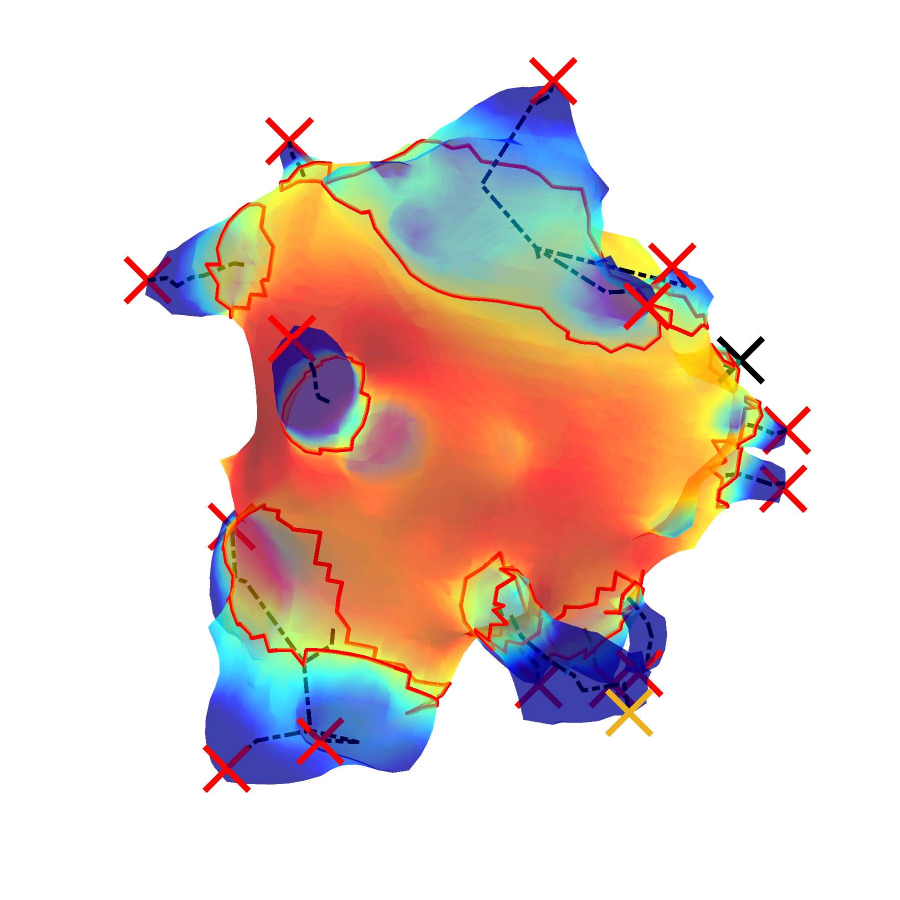}\\
&&&&\\
RS      &   1   &   2   &   3   &   4   &   5   \\
$s_a$   &    5.3&   54.4&   41.9&   68.7&   73.3\\
$s_b$   &    0.3&    0.6&    0.7&    0.9&    0.8\\
$s_1$   &   -0.2&   -0.7&   -0.6&   -1.1&   -1.8\\
$s_2$   &    0.1&   0.2&     0.4&    0.6&    1.6\\
$N_s$   &      0&      1&      4&      8&    14\\
$N_l$   &      2&      0&      3&      1&     1\\
$N_a$   &1&14&4&0&1\\
\end{tabular}\\
\end{center}
\caption{Spiculation quantification and attachment detection via area distortion metric. (a) the results of segmentation and attachment detection on axial slice (blue dashed line: GrowCut segmentation, red dashed line: CIP segmentation, white line: final segmentation, and green region: attachment), (b) 3D shapes of the final segmentation and attached surface (green region: attachment), (c) the results of spiculation detection (red line: baseline of peak, black line: medial axis of peak, red X: spiculation, black X: lobulation, yellow X: attached peaks). Radiologist's spiculation score (RS), Dhara's spiculation scores ($s_a, s_b$\citep{dhara2016differential}), and the proposed interpretable spiculation features ($s_1$: spiculation score, $s_2$: spiculation score, $N_s$: no. of spiculation, $N_l$: no. of lobulation and $N_a$: no. of attached peaks) are shown below. 
\textit{Some small peaks were excluded from the final detection because of the spiculation height threshold.}
\label{fig:segmentation_sphere_param}}
\end{figure*}

\noindent
\textbf{Spiculation measures:} Here, the number of all peaks $N_p$, the number of spiculation $N_s$, the number of lobulation $N_l$, the number of attachment $N_a$, and the surface area ratio of attached regions $r_a=A(S_a)/A(S_{nodule})$ were available as spiculation measures as well as lobulation and attachment measures. We also described novel spiculation scores $s_1$ and $s_2$, $s_1$ summarized sharpness of each spiculation by applying the mean ($\textrm{mean}(\epsilon_{p(i)})$) of the area distortion of all the vertices in a detected spike,
\[s_1 = \frac{\sum_{i}{\textrm{mean}(\epsilon_{p(i)})*h_{p(i)}}}{\sum_{i}{h_{p(i)}}}\] where $p(i)$ is spike $i$, $h_{p(i)}$ is height of spike $p(i)$, and
$s_2$ summarized irregularity of spiculation by applying the variation ($\textrm{var}(\epsilon_{p(i)})$) of the area distortion,
\[s_2 = \frac{\sum_{i}{\textrm{var}(\epsilon_{p(i)})*h_{p(i)}}}{\sum_{i}{h_{p(i)}}}.\]

Fig.~\ref{fig:segmentation_sphere_param} illustrates a few nodule examples and the corresponding spiculation quantification measures and radiologist's spiculation scores.

We compared our spiculation measures with \cite{dhara2016differential} proposed spiculation scores $s_a$ and $s_b$,
\[s_a=\sum_{i}{e^{-\omega_{p(i)}}h_{p(i)}}\] and
\[s_b=\frac{\sum_{i}{h_{p(i)}\cos{\omega_{p(i)}}}}{\sum_{i}{h_{p(i)}}},\]
where $\omega_{p(i)}$ is the solid angle subtended at apex of spike $p(i)$. These scores summarized the sharpness (solid angle $\omega_{p(i)}$) and height ($h_{p(i)}$) of spiculations as shown in Figure \ref{fig:segmentation_sphere_param}.

\subsection{Semi-automatic segmentation} \label{sec:segmentation}



Spiculations are thin sharp spikes around the core of a nodule. We used semi-automatic segmentation to precisely segment/quantify spiculations as well as extract radiomic features. A consensus contour was generated for each nodule with two or more manual contours by using the simultaneous truth and performance level estimation (STAPLE)~\citep{warfield2004simultaneous,choi2016individually} as ground truth. Many automatic segmentation methods have been proposed for nodule segmentation, but are mainly focused on segmenting core nodule regions. Furthermore, the implementation of these segmentation methods is not straightforward.

\begin{figure}[h]
\begin{center}
\setlength{\tabcolsep}{1pt}
\begin{tabular}{ccccc}
\includegraphics[width=0.12\textwidth]{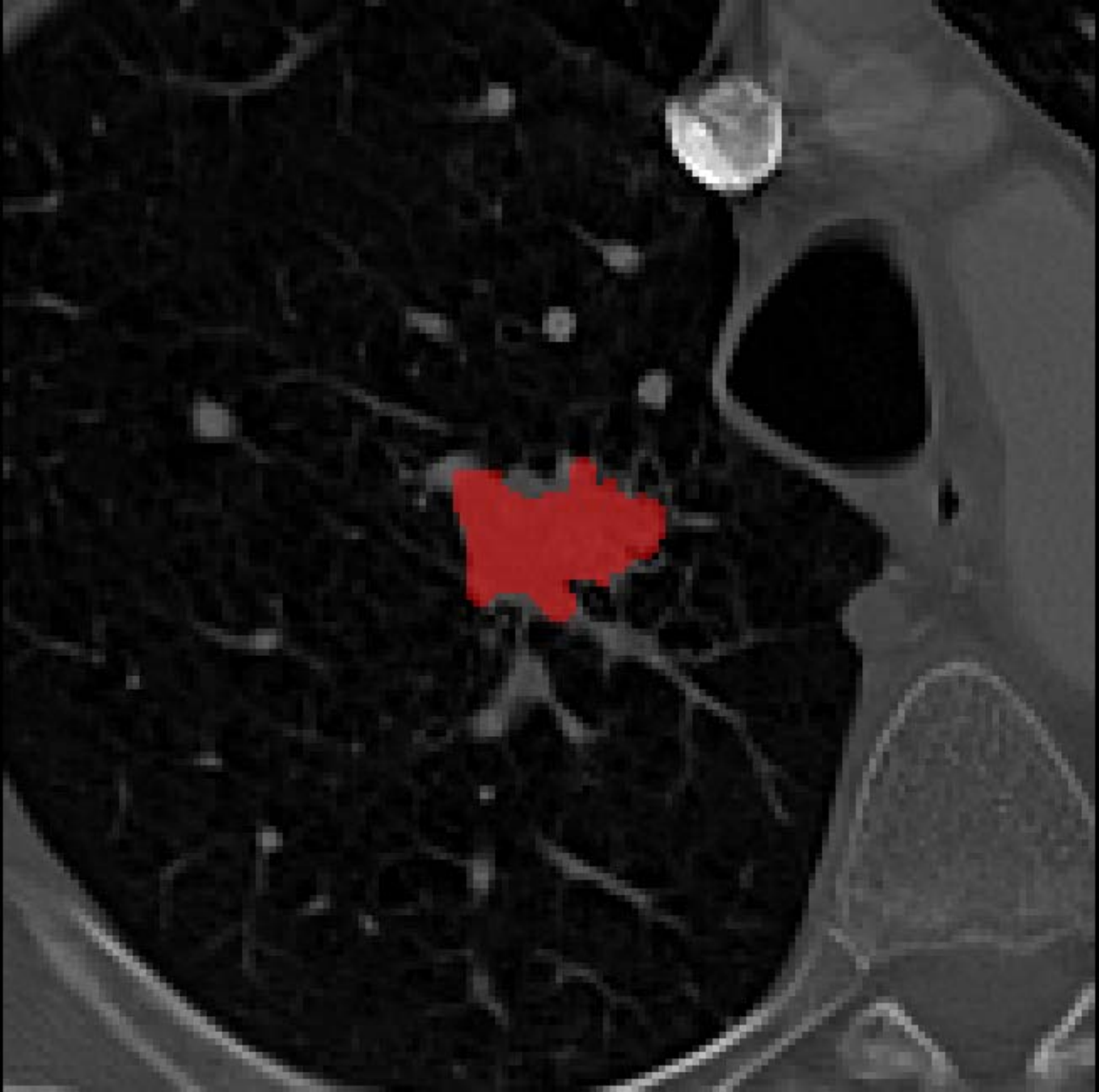}&
\includegraphics[width=0.12\textwidth]{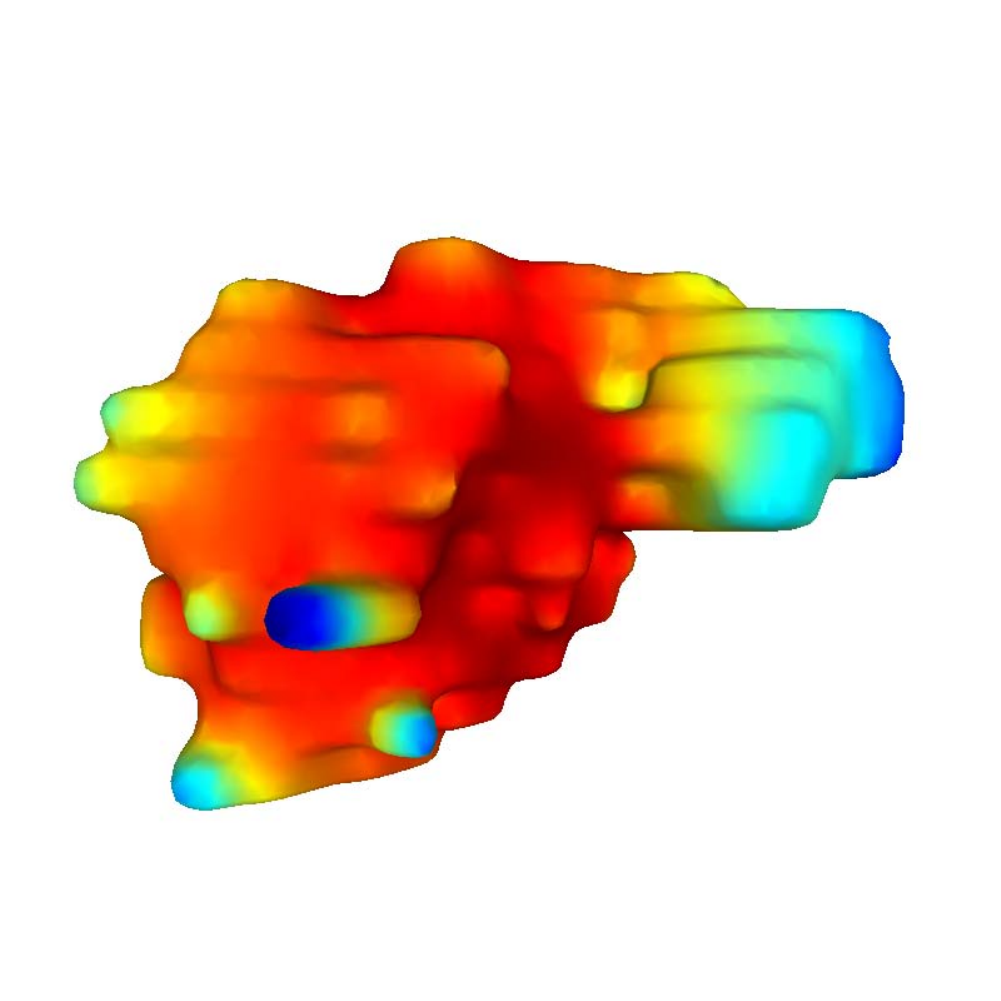}&
\includegraphics[width=0.12\textwidth]{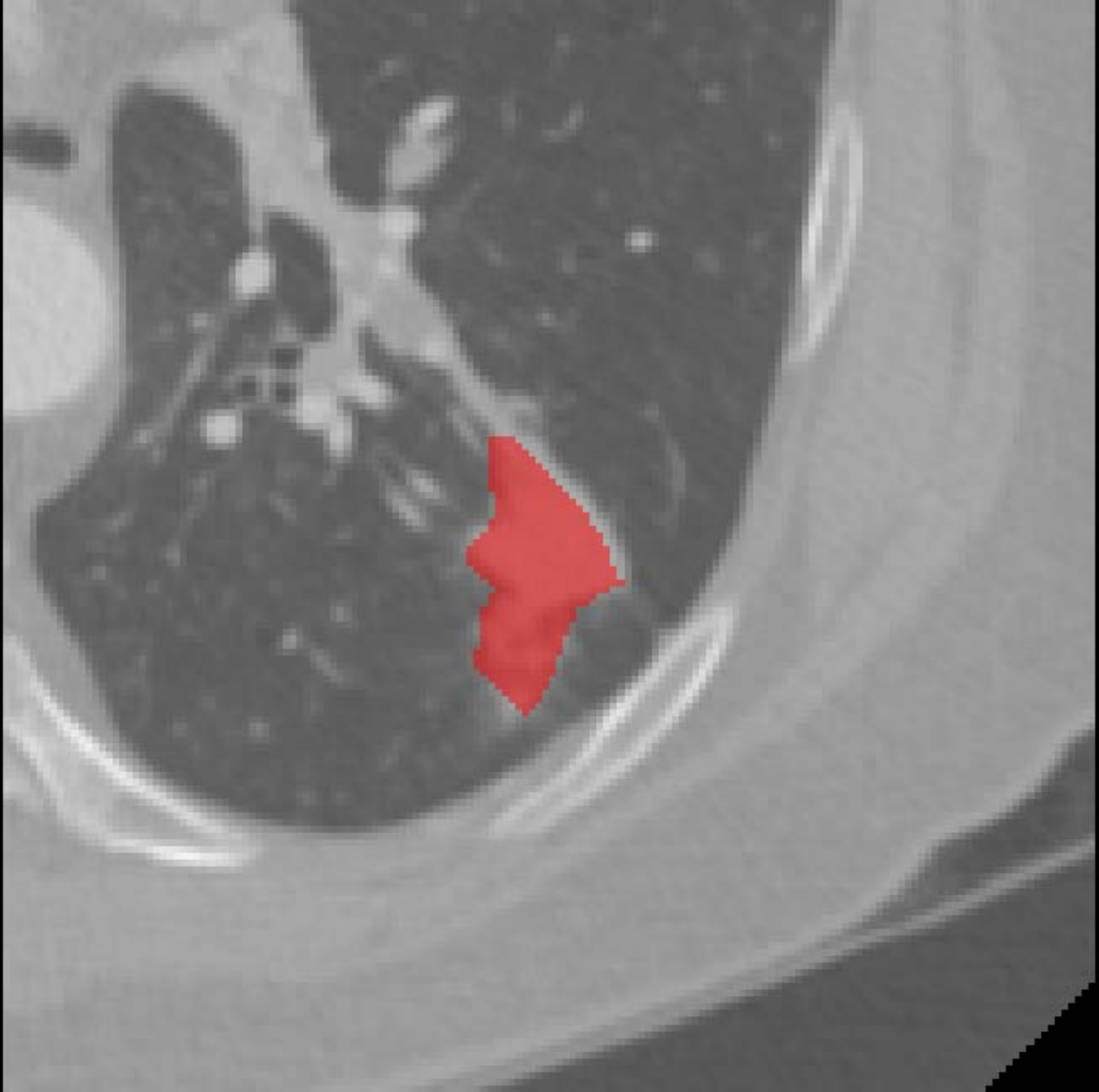}&
\includegraphics[width=0.12\textwidth]{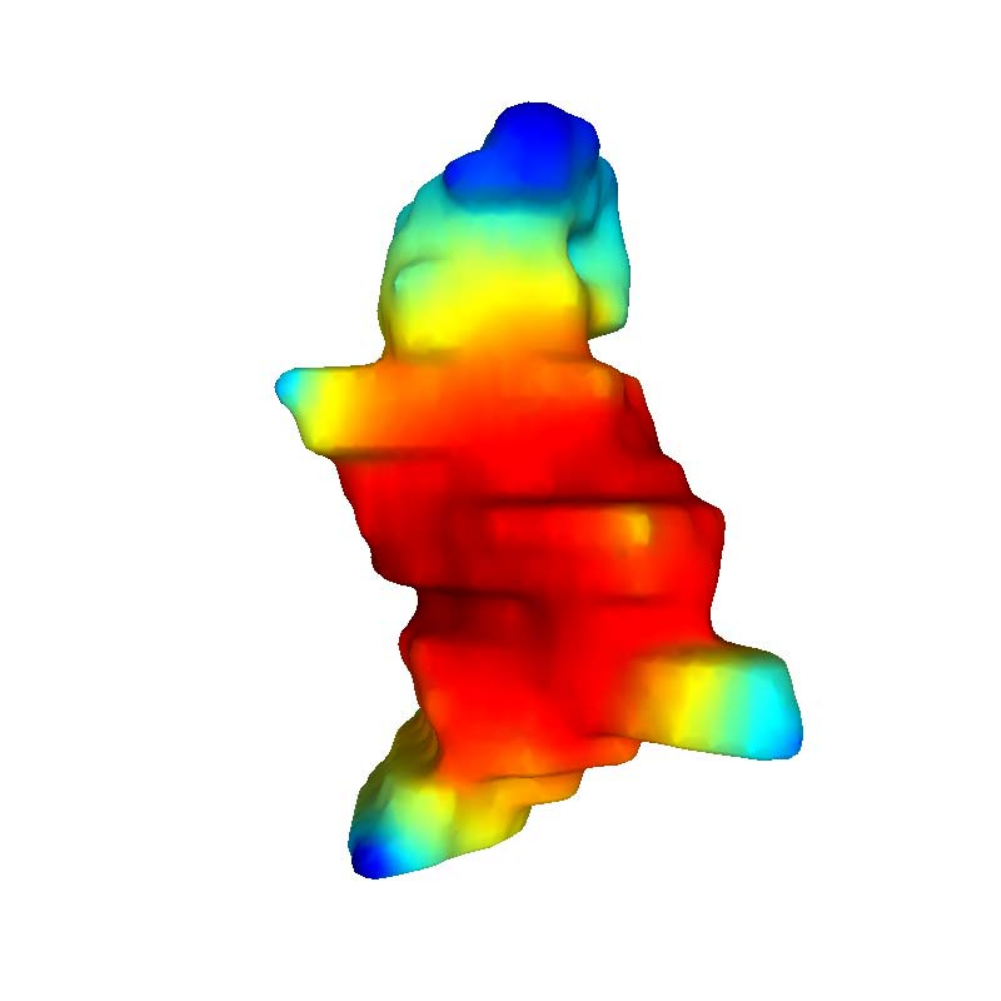}\\
\includegraphics[width=0.12\textwidth]{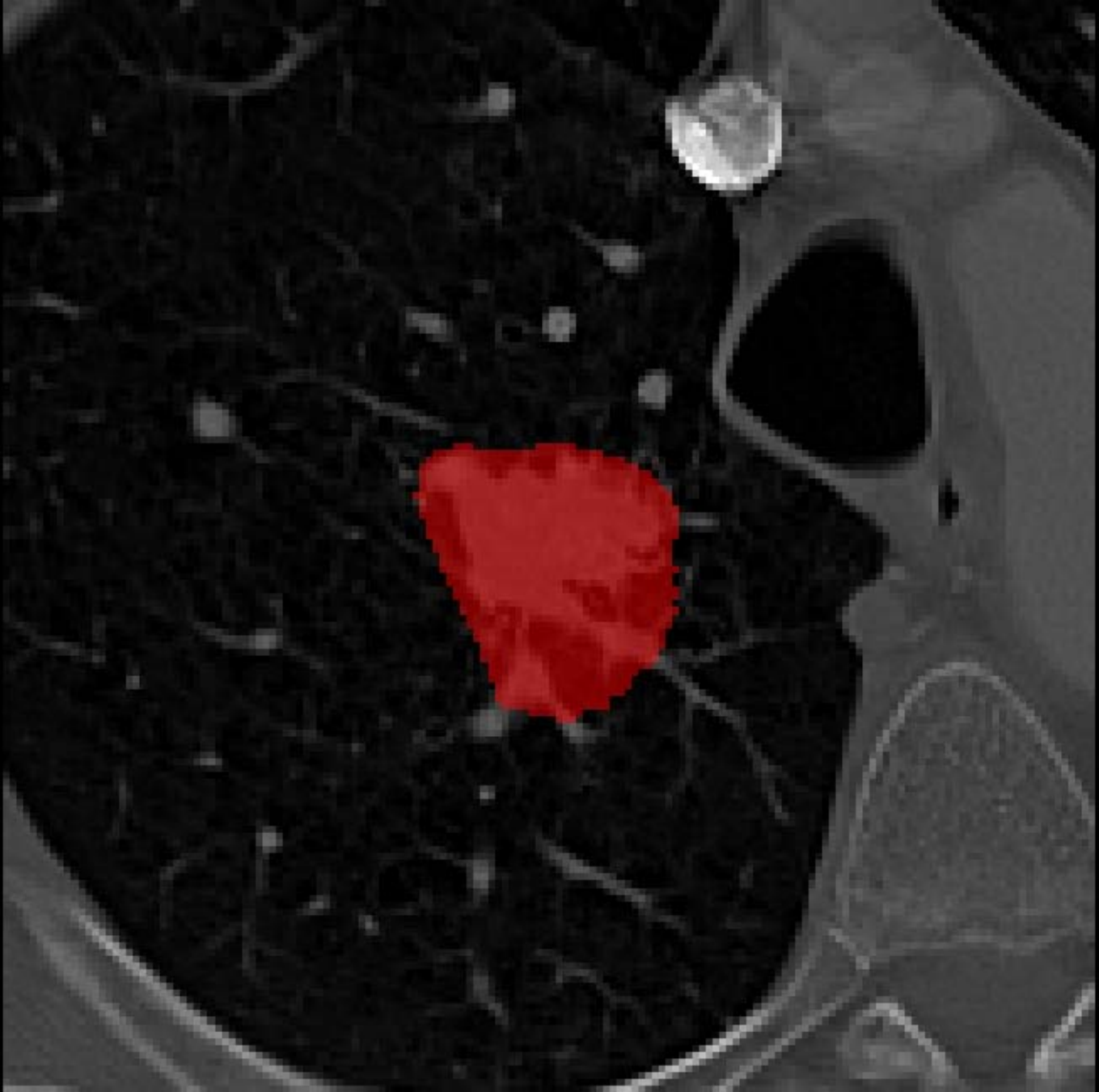}&
\includegraphics[width=0.12\textwidth]{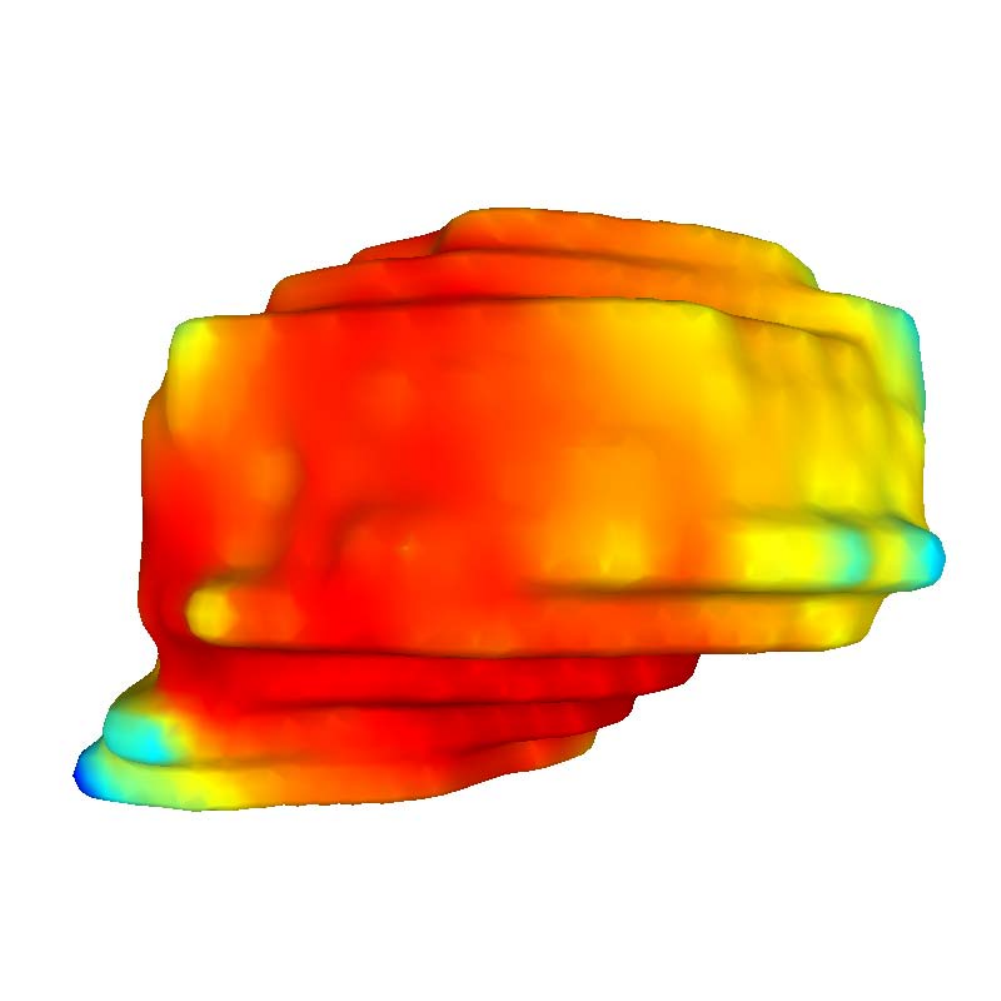}&
\includegraphics[width=0.12\textwidth]{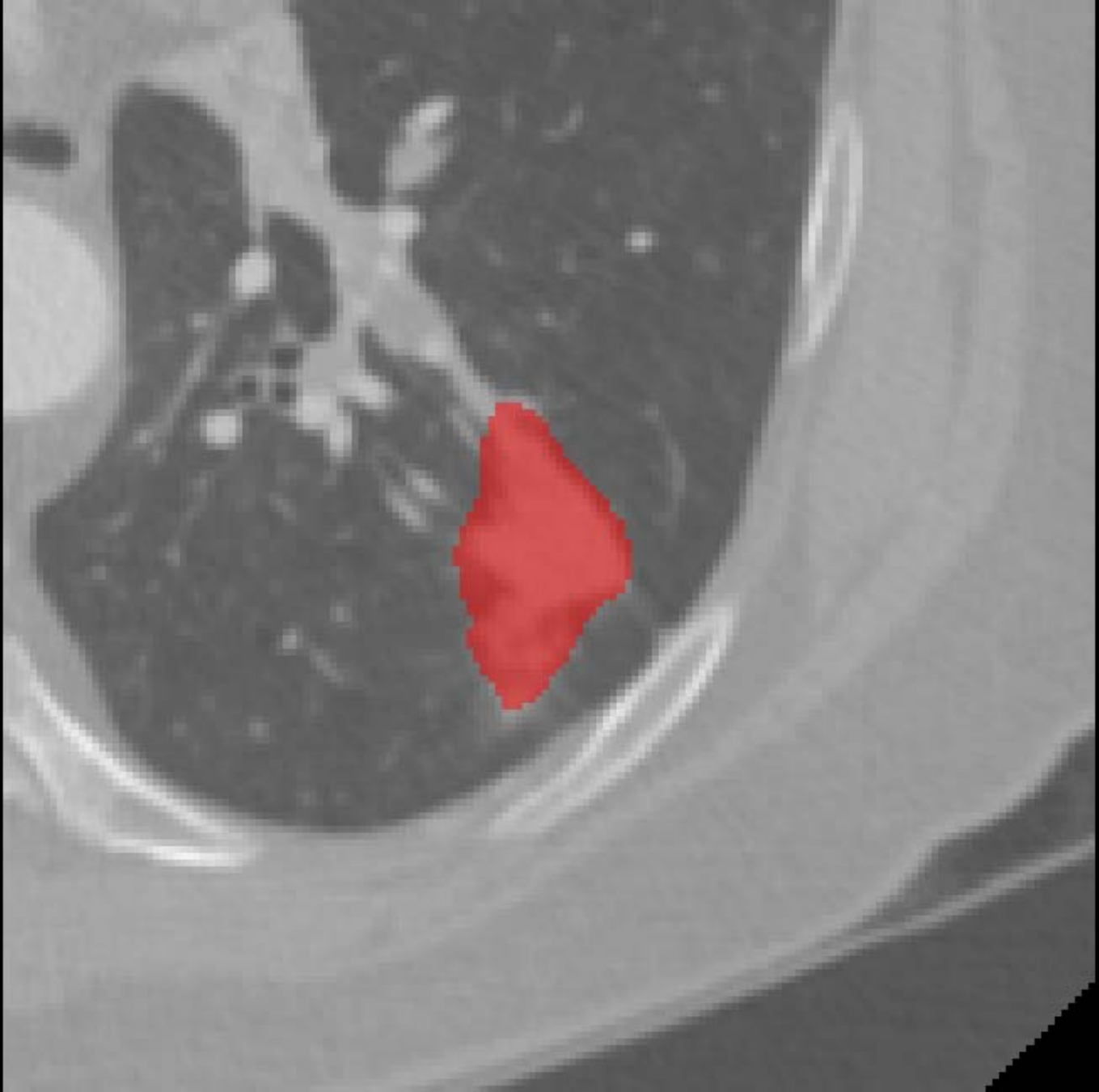}&
\includegraphics[width=0.12\textwidth]{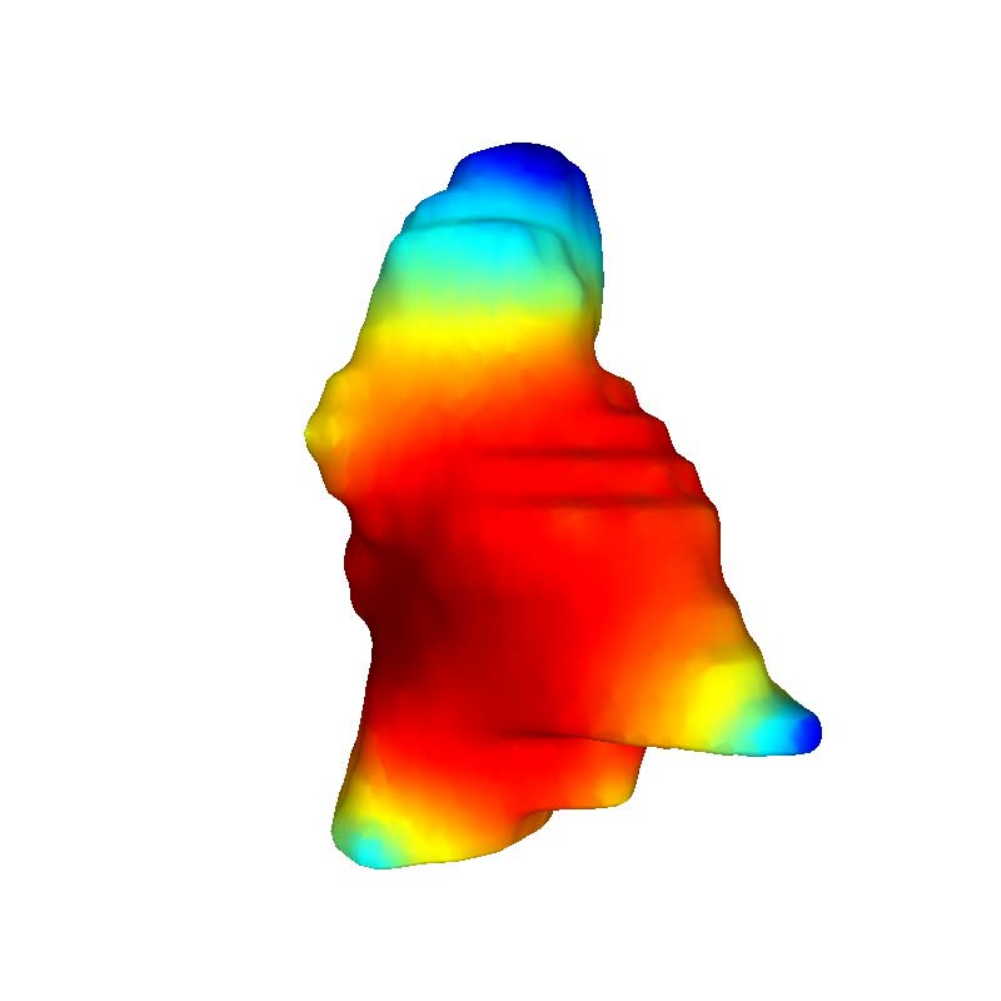}\\
\end{tabular}
\end{center}
\caption{First row illustrates two LIDC cases with ground truth manual contours where the radiologists meticulously marks the spiculations as per the Lung-RADS guidelines. In contrast, deep learning algorithms \cite{jiang2018multiple,jiang2020psigan} though accurate in localization can completely smooth out the critical spiculation features. Our proposed combination gives more reproducible segmentation results and is designed to preserve the spiculations as best as possible.
\label{fig:dl_limitation}}
\end{figure}

For the reproducible semi-automatic segmentation, we combined two well-known and easy-to-implement methods, GrowCut \citep{vezhnevets2005growcut} and chest imaging platform (CIP) segmentation algorithms \citep{yip2017application}. The GrowCut Segmentation is a cellular automata-based region growing algorithm that needs two sets of seed points for foreground and background, and they compete to grow the regions until convergence. GrowCut segmentation can leak into surrounding structures, such as the chest wall, airway walls, and vessel-like structures. The CIP segmentation is a level set-based algorithm that uses a front propagation approach from a seed point placed within the nodule. The propagation (or segmentation) is constrained by feature maps of the structures to prevent leakage into surrounding structures. However, CIP might ignore some tumor regions because of the inaccurate vessel and wall feature maps. Therefore, we combined these two methods to compensate for their limitations and to take advantage of both for attachment detection. Both methods are publicly available in 3D Slicer. Fig.~\ref{fig:segmentation_sphere_param}(a) show segmentations of different nodules. The corresponding attachments, shown in Fig. \ref{fig:segmentation_sphere_param}(a) and (b), are computed using the morphological intersection of the GrowCut and CIP segmentations.

\subsection{Malignancy Prediction} \label{sec:malignancy}
We evaluated our spiculation quantification measures and radiomic features for classifying pathological malignant nodules and benign nodules by adding spiculation features to our model~\citep{choi2018medphy}. The conventional spiculation scores ($s_a$ and $s_b$) were also evaluated. A total of 103 radiomic features was extracted from each nodule to quantify its intensity, shape, and texture~\cite{choi2018medphy}. Intensity features are first‐order statistical measures that quantify the level and distribution of CT attenuations in a nodule (e.g., Minimum, Mean, Median, Maximum, Standard deviation (SD), Skewness, and Kurtosis). Shape features describe geometric characteristics (e.g., volume, diameter, elongation, roundness, and flatness) for voxels. Texture features quantify tissue density patterns. We used Gray‐level co‐occurrence matrix (GLCM): Energy, Entropy, Correlation, Inertia, Cluster prominence (CP), Cluster shade (CS), Haralick’s correlation (HC), Inverse difference moment (IDM); and Gray‐level run‐length matrix (GLRM): Run-length non-uniformity (RNU), Gray-level non-uniformity (GNU), Long-run emphasis (LRE), Short-run emphasis (SRE), High gray-level run emphasis (HGRE), Low gray-level run emphasis (LGRE), Long-run high gray-level emphasis (LRHGE), Long-run low gray-level emphasis (LRLGE), Short-run high gray-level emphasis (SRHGE), Short-run low gray-level emphasis (SRLGE). The mean (average) and SD values of each texture feature were computed over 13 directions to obtain rotationally invariant features.

Moreover, we extracted features from the triangular mesh model, such as shape features (size - volume, average of longest and its perpendicular diameters, equivalent volume sphere's diameter, and roundness) and statistical features (median, mean, minimum, maximum, variance, skewness, and kurtosis) of the area distortion metric $\epsilon$. We performed univariate analysis to evaluate the significance of each feature to classify spiculation using the area under the receiver operating characteristic curve (AUC), Wilcoxon rank-sum test, and Spearman's correlation coefficient $\rho$. Bonferroni correction was applied to the original $p$-values to counteract the problem of multiple comparisons since the multiple features were tested for a single outcome.

We applied the SVM-LASSO model~\citep{choi2018medphy} to predict the malignancy of nodules. The model uses size (BB\_AP) and texture (SD\_IDM) features. We inter-compared the original SVM-LASSO model and other feature combinations of the two features and new spiculation features or radiologist's spiculation score (RS), respectively. We use the same data set and evaluation method in \cite{choi2018medphy} to evaluate the new radiomics model with spiculation. Moreover, we also evaluated another model building process using weak-labeled data (radiological malignancy score, RM) to predict pathological malignancy (PM), which allowed more data to be used despite missing pathological malignancy.

\begin{figure}[h]
\begin{center}
\includegraphics[width=0.45\textwidth]{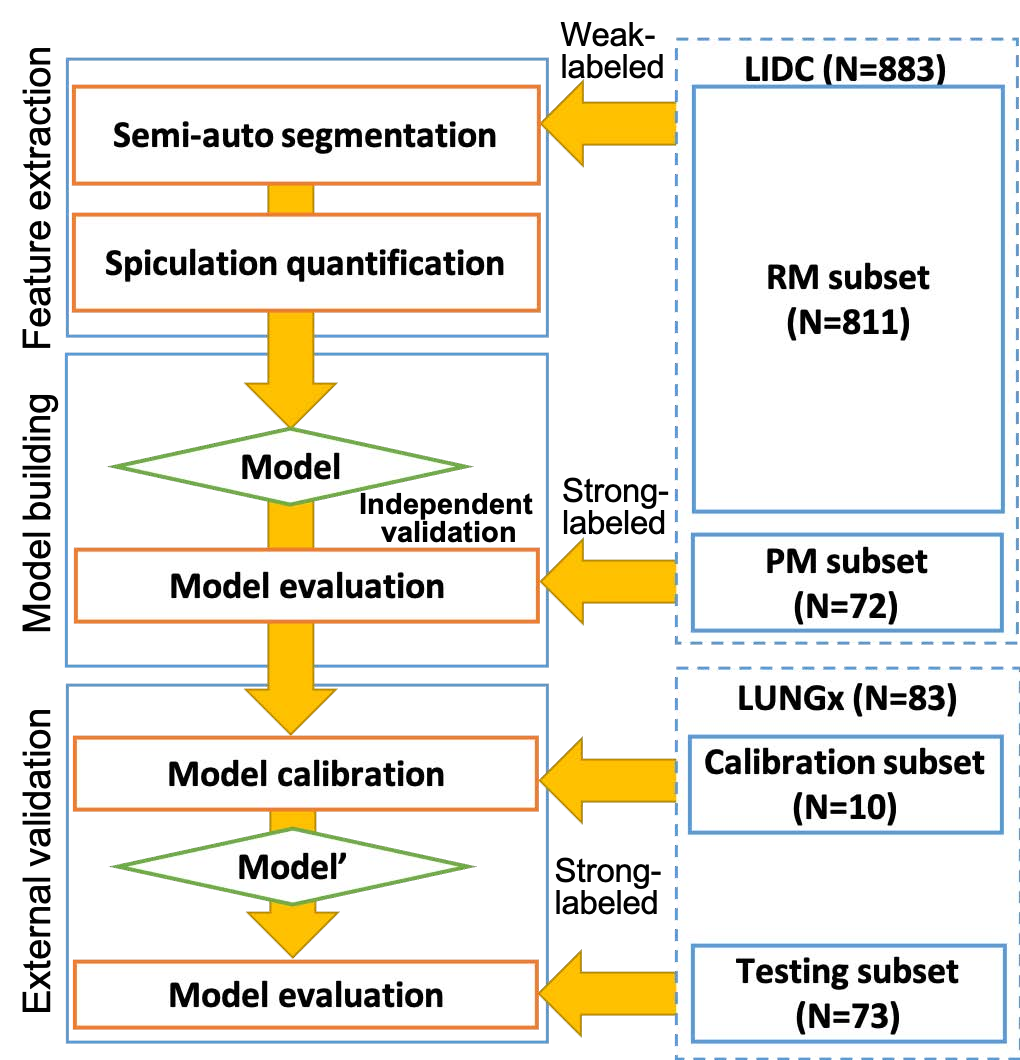}\\
\end{center}
\caption{The data flow of the malignancy prediction model building and external validation. We applied semi-auto segmentation to nodules in both LIDC and LUNGx datasets, and features were extracted for these segmentations. The extracted features from LIDC dataset were used for model building. The model were evaluated on pathological malignancy subset and the LUNGx test set after model calibration using the LUNGx calibration set (Model').
\label{fig:data_flow}}
\end{figure}

\section{Results} \label{sec:results}
We have evaluated the proposed spiculation quantification method by comparing with radiologists score and applied the interpretable spiculation features in the nodule malignancy prediction, which is the main task in the lung cancer screening.
\subsection{Data Preparation} \label{sec:results:data}
The Lung Image Database Consortium image collection (LIDC-IDRI) \citep{armato2015lidc,armato2011lidc} and LUNGx datasets \citep{armato2015spie,armato2016lungx} were applied to evaluate the proposed method, and the data flow is shown in Fig.~\ref{fig:data_flow}. LIDC contains 1018 cases with low-dose screening thoracic CT scans and marked-up annotated lesions. Four experienced thoracic radiologists annotated nodules, including delineation, malignancy (RM), spiculation (RS), margin, texture, and lobulation. Eight hundred eighty-three cases in the dataset have nodules with contours. For the biggest nodules in each case, we applied semi-auto segmentation for more reproducible spiculation quantification and also calculated consensus segmentation using STAPLE to combine multiple contours by the radiologists. The accuracy of our semi auto-segmentation compared to the consensus contour was 0.71$\pm$0.13 in terms of the dice coefficient. LUNGx consists of 10 cases for calibration set (10 nodules) and 60 cases for the test set (73 nodules). We applied the same semi-auto segmentation to nodules in the LUNGx dataset.

\label{radiologists_score}For more rigorous data analysis, we divided the LIDC dataset into two subsets depending on whether pathological malignancy (LIDC\_PM, N=72) or radiological malignancy (LIDC\_RM, N=811) was available. The radiological malignancy scores are 1 - highly unlikely, 2 - moderately unlikely, 3 - indeterminate likelihood, 4 - moderately suspicious, and 5 - highly suspicious for cancer. RM$>$3 (moderately suspicious to highly suspicious) was considered radiological malignancy. RS in the dataset ranged between 1 (non-spiculated) and 5 (highly spiculated). There are up to four annotations for each nodule, we aggregated the scores by applying voting. When there are two most frequent scores, we chose higher score. \textit{We binarized the RS using three different cutoffs (1,2, and 3) because the current clinical standard uses binary classification, non-spiculated (NS) and spiculated (S)}, as shown in Table~\ref{table:balance} (the cutoff at four is shown for reference).

\begin{table}[t]
\centering
\caption{The numbers of the non-spiculated (NS, $\leq T_s$) and spiculated (S, $> T_s$) nodules for the subsets, LIDC\_PM $(N=72)$ and LIDC\_RM $(N=811)$.}
\label{table:balance}
\setlength{\tabcolsep}{6pt}
\begin{tabular}{rcc|cc}
\hline
 & \multicolumn{2}{c|}{LIDC\_PM} & \multicolumn{2}{c}{LIDC\_RM}\\
$T_s$ & NS & S & NS & S\\
\hline
1 & 35 & 37 & 474 & 337 \\
2 & 55 & 17 & 704 & 107 \\
3 & 58 & 14 & 747 &  64 \\
4 & 67 &  5 & 790 &  21 \\
\hline
\end{tabular}
\end{table}

\begin{figure}[h]
\begin{center}
\includegraphics[height=0.3\textwidth,width=0.4\textwidth]{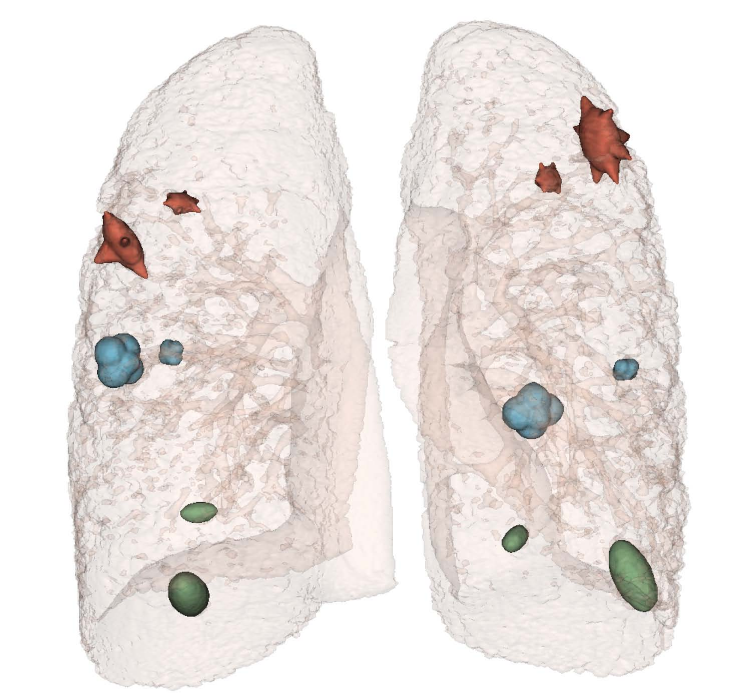}\\
\end{center}
\caption{Nodules on Phantom FDA layout \#4 were used for optimizing spiculation height and solid angle thresholds ($T_h$ and $T_\Omega$) to remove false positives in spiculation detection.
\label{fig:phantom_FDA_4}}
\end{figure}

\begin{figure*}[h]
\begin{center}
\setlength{\tabcolsep}{3pt}
\begin{tabular}{cm{0.15\textwidth}m{0.15\textwidth}m{0.15\textwidth}m{0.15\textwidth}m{0.5cm}c}
&\multicolumn{2}{c}{Left Lung}&\multicolumn{2}{c}{Right Lung}\\
(a)&
\includegraphics[width=0.15\textwidth]{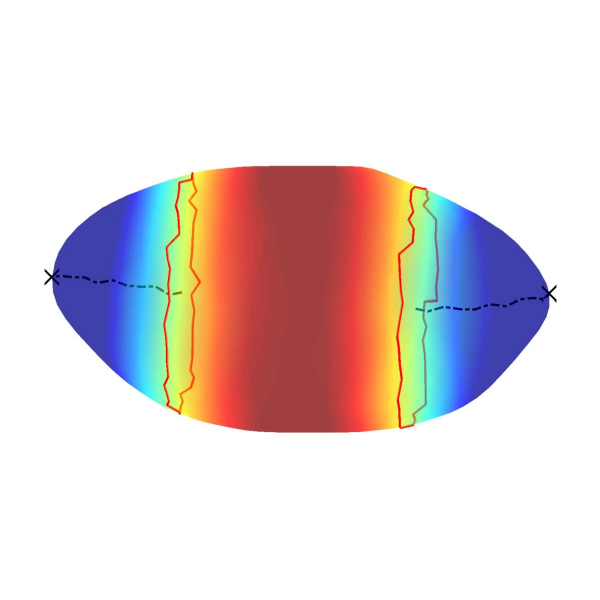}&
\includegraphics[width=0.15\textwidth]{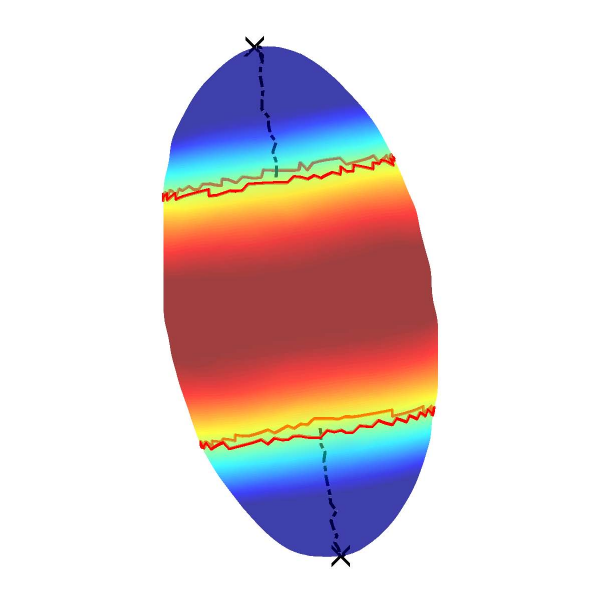}&
\includegraphics[width=0.15\textwidth]{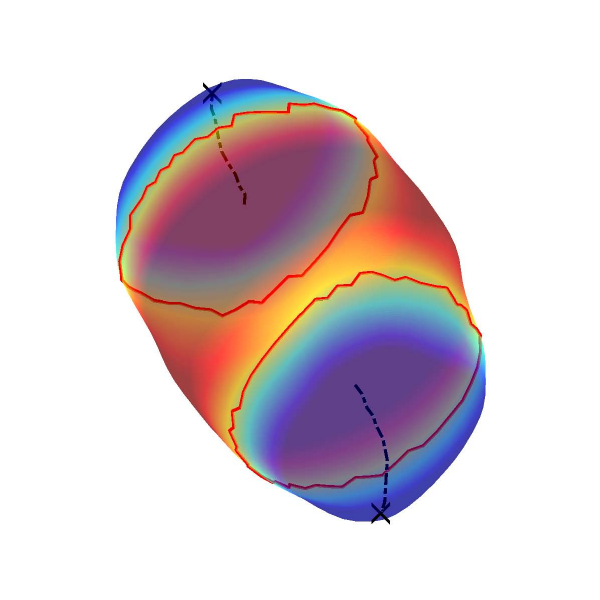}&
\includegraphics[width=0.15\textwidth]{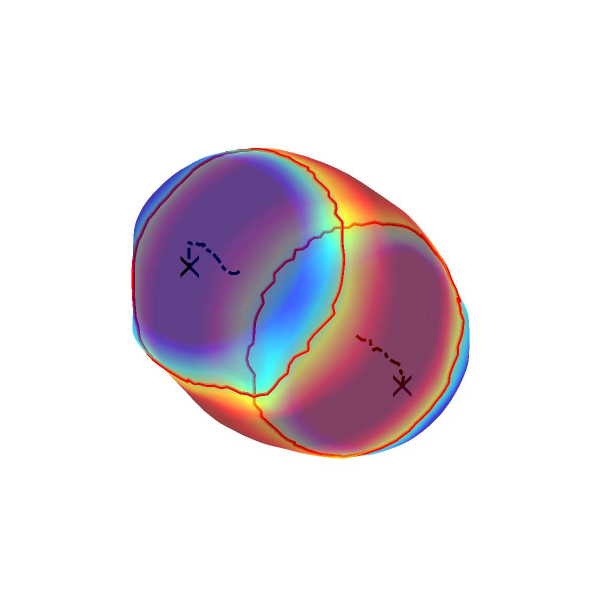}&
\includegraphics[height=0.12\textwidth]{disp2}\\
(b)&
\includegraphics[width=0.15\textwidth]{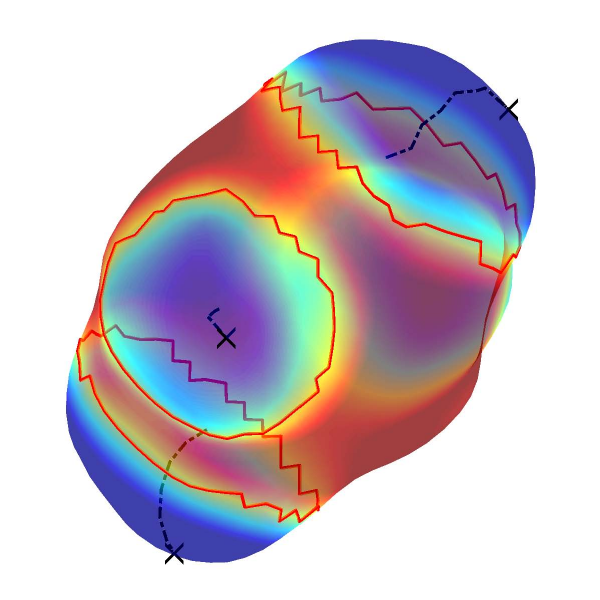}&
\includegraphics[width=0.15\textwidth]{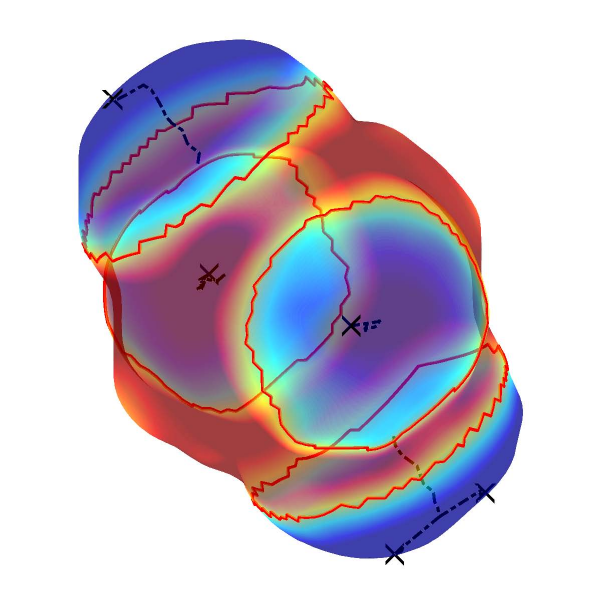}&
\includegraphics[width=0.15\textwidth]{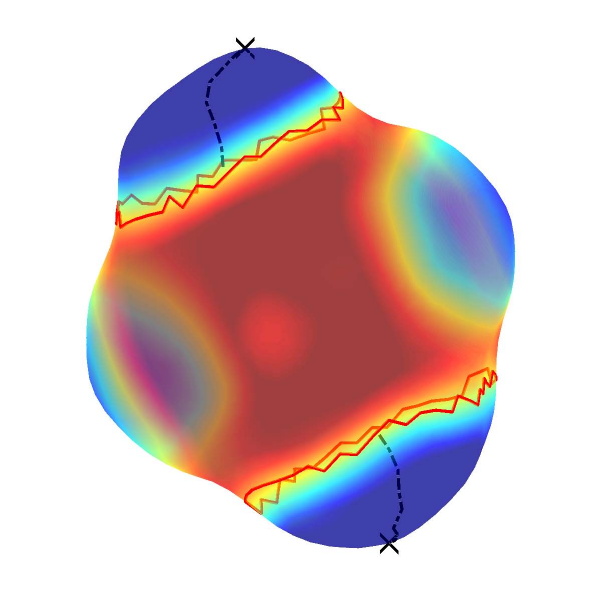}&
\includegraphics[width=0.15\textwidth]{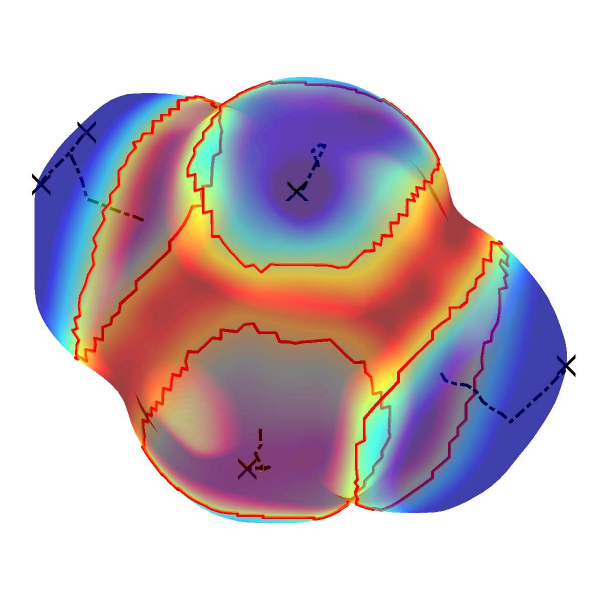}&
\includegraphics[height=0.12\textwidth]{disp2}\\
(c)&
\includegraphics[width=0.15\textwidth]{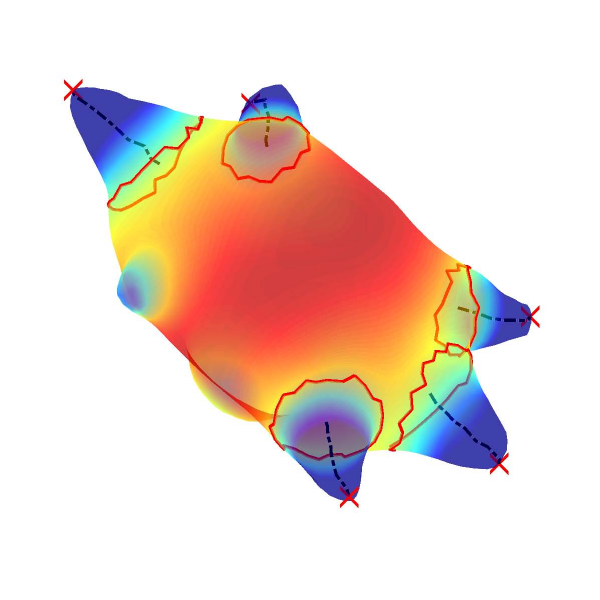}&
\includegraphics[width=0.15\textwidth]{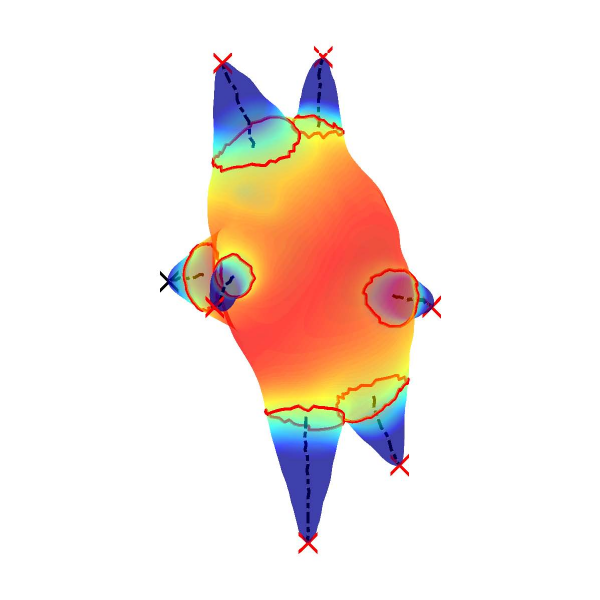}&
\includegraphics[width=0.15\textwidth]{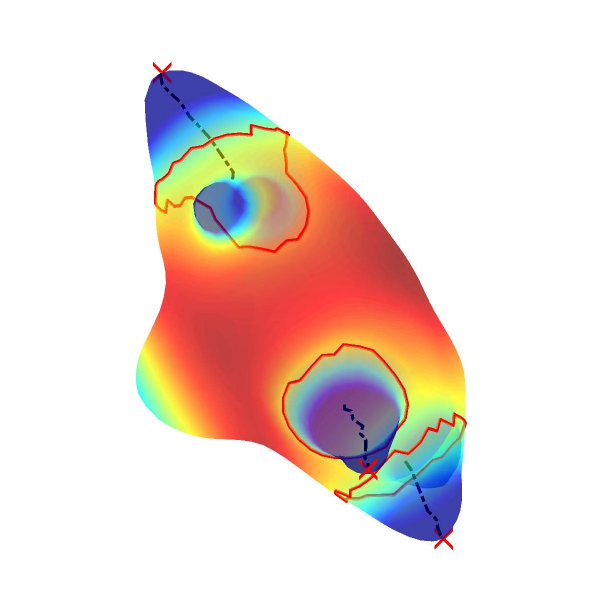}&
\includegraphics[width=0.15\textwidth]{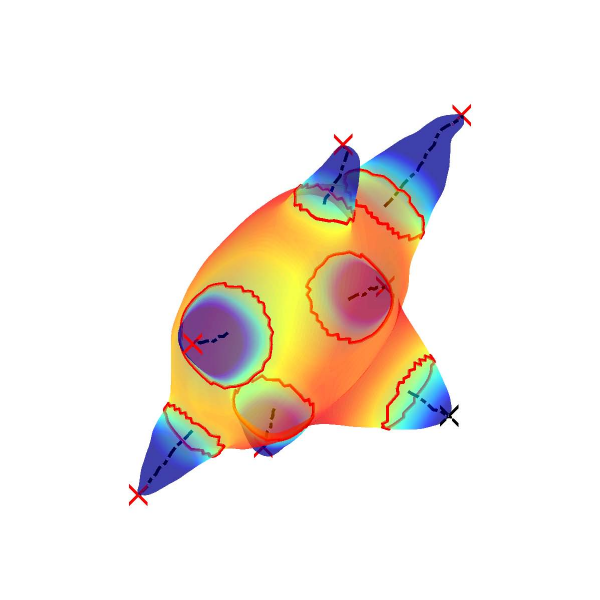}&
\includegraphics[height=0.12\textwidth]{disp2}\\
\end{tabular}
\end{center}
\caption{Spiculation quantification results of Phantom FDA layout \#4, (a) elliptical, (b) lobulated and (c) spiculated (red line: baseline of peak, black line: medial axis of peak, red X: spiculation, black X: lobulation, first and third columns: 10mm, second and fourth columns: 20mm).
\label{fig:sphere_param_examples}}
\end{figure*}

To optimize spiculation height and solid angle thresholds, $T_h$ and $T_\Omega$, for filtering out false positives such as small peaks and lobulations, we used Phantom FDA layout \#4 as shown in Fig.~\ref{fig:phantom_FDA_4} \citep{gavrielides2015phantomfda,gavrielides2010phantomfda,clark2013tcia}. We tuned the thresholds to clearly differentiate the spiculations from lobulations (annotations available in Phantom FDA) and to detect as many spiculations as possible without false positives. The final selected thresholds were $T_h\ge3 mm$ and $T_\Omega\le0.65 sr$. Fig.~\ref{fig:sphere_param_examples} shows the results of spiculation quantification for each nodule in the phantom data. The optimal thresholds excluded lobulations and elliptical shape corners from final spiculations.

\subsection{Spiculation Quantification} \label{sec:results:spiculation}

The proposed method was implemented in Matlab 2017b, and experiments were performed on a workstation with Intel(R) Xeon(R) CPU E5-1620 v2 @ 3.70GHz and 32 GB RAM running macOS 10.15.5. Table.~\ref{table:runtime} shows average run-time of the spiculation quantification and its major components.
 
\begin{table}[]
\centering
\caption{The run-time of spiculation quantification for each nodule.}
\label{table:runtime}
\setlength{\tabcolsep}{6pt}
\begin{tabular}{lr}
& Average Run-time (s) \\
\hline
 Algorithm~\ref{algo:peak_detection} & 0.29 \\
 Curvature Computation & 0.65 \\
 Other Computations & 0.42   \\
\hline
 Total &  1.36 \\

\end{tabular}
\end{table}

\begin{table*}[h]
\centering
\caption{Twenty highly correlated features with radiologist's spiculation score in univariate analysis. $T_s$: the threshold to binarize the radiologist's spiculation score}
\vspace{3mm}
\label{table:good_features_seg}
\setlength{\tabcolsep}{3pt}
\begin{tabular}{rlccccr}
\hline
Rank & Feature name & \multicolumn{4}{c}{AUC} & Corr\\
 &  & $T_s>1$ & $T_s>2$ & $T_s>3$ & Average & $\rho$\\
\hline
1  &                $N_s$ &  0.73 &  0.79 &  0.81 &  0.78 &  0.44 \\
2  &      Roundness(mesh) &  0.73 &  0.83 &  0.82 &  0.79 & -0.44 \\
3  &               SD LRE &  0.74 &  0.80 &  0.76 &  0.77 & -0.44 \\
4  &      Mean $\epsilon$ &  0.72 &  0.80 &  0.81 &  0.77 & -0.41 \\
5  &   Minimum $\epsilon$ &  0.72 &  0.79 &  0.79 &  0.77 & -0.41 \\
6  &           Mean LRLGE &  0.72 &  0.76 &  0.75 &  0.74 & -0.40 \\
7  &             SD LRLGE &  0.71 &  0.77 &  0.77 &  0.75 & -0.39 \\
8  & Weighted Principal Moment 1 &  0.71 &  0.76 &  0.78 &  0.75 &  0.39 \\
9  &    Median $\epsilon$ &  0.71 &  0.79 &  0.79 &  0.76 & -0.39 \\
10 &  Variance $\epsilon$ &  0.70 &  0.78 &  0.79 &  0.76 &  0.39 \\
11 &   Maximum $\epsilon$ &  0.71 &  0.77 &  0.78 &  0.75 &  0.38 \\
12 &                SD CS &  0.71 &  0.73 &  0.71 &  0.72 & -0.38 \\
13 &2D Weighted Principal Moment 1 &  0.71 &  0.74 &  0.76 &  0.73 &  0.37 \\
14 &                $s_1$ &  0.70 &  0.77 &  0.79 &  0.75 & -0.37 \\
15 &              SD LGRE &  0.70 &  0.73 &  0.73 &  0.72 & -0.36 \\
16 &  2D Roundness(voxel) &  0.68 &  0.78 &  0.77 &  0.74 & -0.34 \\
17 &                $s_2$ &  0.68 &  0.75 &  0.77 &  0.73 &  0.34 \\
18 &             SD SRLGE &  0.68 &  0.72 &  0.71 &  0.70 & -0.33 \\
19 &          Mean Energy &  0.69 &  0.69 &  0.66 &  0.68 & -0.33 \\
20 &               2D Sum &  0.67 &  0.72 &  0.73 &  0.71 & -0.32 \\
\hline
\end{tabular}
\end{table*}

Since we evaluated the malignancy prediction model on LIDC\_PM, we performed univariate analysis on LIDC\_RM to avoid the selection bias in malignancy prediction model building. In the univariate analysis, 84 features were identified as significant features (adjusted $p$-value$<$0.05) for spiculation quantification. Among these, 56 features were highly correlated with size features ($\rho > 0.75$); size is one of the main criteria for diagnosing malignancy. Thus, we removed all the size-related features, including $s_a$ and $N_p$, to provide complementary information. After applying the size-related feature removal, 28 significant features remained, and we picked 20 highly correlated features with RS. Half of these were texture or intensity statistics features, which are not interpretable. Almost all of our spiculation measures were significant and ranked in the top 20. Table~\ref{table:good_features_seg} show the univariate analysis results of the top 20 features using semi-auto segmentation. None of Dhara's spiculation scores ($s_a$ and $s_b$) were selected in the top 20 features even though they were significant features. $s_a$ was excluded by its high correlation with size ($\rho=0.87$), and $s_b$ was not ranked among the top 20.

\begin{table*}[]
\centering
\caption{Spiculation classification results using the interpretable spiculation features, $N_s$, $N_a$, $N_l$, $N_p$, $r_a$, $s_1$, and $s_2$. $T_s$: the threshold to binarize the radiologist's spiculation score}
\label{table:spiculation_results}
\setlength{\tabcolsep}{6pt}
\begin{tabular}{rlcccc}
\hline
&Threshold & Sensitivity & Specificity & Accuracy & AUC  \\
\hline
\multicolumn{6}{l}{\textit{10x10-fold CV on LIDC\_PM}}\\
&$T_s>1$ & 69.9$\pm$2.1\%     & 87.5$\pm$1.8\%     & 79.7$\pm$1.7\%  & 0.84$\pm$0.01 \\
&$T_s>2$ & 83.8$\pm$3.9\%     & 83.9$\pm$2.7\%     & 83.9$\pm$2.4\%  & 0.90$\pm$0.01 \\
&$T_s>3$ & 73.0$\pm$2.9\%     & 81.7$\pm$1.2\%     & 79.7$\pm$0.9\%  & 0.87$\pm$0.01 \\
\hline
\multicolumn{6}{l}{\textit{10x10-fold CV on LIDC\_RM}}\\
&$T_s>1$ & 47.9$\pm$0.4\%     & 85.8$\pm$0.3\%     & 69.7$\pm$0.3\%  & 0.75$\pm$0.01 \\
&$T_s>2$ & 73.2$\pm$0.4\%     & 79.7$\pm$0.4\%     & 78.8$\pm$0.3\%  & 0.81$\pm$0.01 \\
&$T_s>3$ & 75.2$\pm$2.0\%     & 79.4$\pm$0.3\%     & 79.1$\pm$0.4\%  & 0.83$\pm$0.01 \\
\hline
\end{tabular}
\end{table*}

We also performed multivariate analysis to classify PNs into spiculated or non-spiculated on different thresholds ($T_s$). The multivariate classification models were evaluated by 10 times 10-fold cross validation (CV) The classification performance is shown in Table \ref{table:spiculation_results}. Highly spiculated PNs ($T_s>2$ and $T_s>3$) were accurately classified in both subsets (LIDC\_PM: accuracy=83.9\%, 79.7\% and LIDC\_RM: accuracy=78.8\%, 79.1\%), but all the spiculated PNs ($T_s>1$) were not stable in the classification accuracy (LIDC\_PM: accuracy=79.7\% and LIDC\_PM: accuracy=69.7\%).

\subsection{Malignancy Prediction} \label{sec:results:malignancy}

We built models using feature combinations of the previously selected features (Size: BB\_AP and Texture: SD\_IDM) from \cite{choi2018medphy}, and the interpretable spiculation features ($N_s,N_a,N_l,N_p,r_a,s_1,s_2$). As shown in Fig. \ref{fig:data_flow}, the model trained by LIDC was then externally validated by LUNGx dataset, which was collected for a lung cancer screening competition (LUNGx Challenge) and provides a calibration set (size-matched ten nodules, five benign and five malignant) and a test set (73 nodules, 37 benign and 36 malignant) \citep{armato2015spie,armato2016lungx,clark2013tcia}. For the external validation, we followed the model evaluation process of the LUNGx Challenge \cite{armato2016lungx}. The model was calibrated by the calibration set of LUNGx (Model') and finally evaluated by the test set (73 cases) of LUNGx. We used zero value instead of missing variable RS in the external validation because LUNGx does not provide it. Since pathological malignancy (PM) was only available for the 72 cases, we used weak-labeled data (LIDC\_RM, N=811) based on the radiological malignancy score (RM). We divided the weak-labeled data into two groups (training 80\% and validation 20\%) for training and optimizing the model. Then, the best model was evaluated on strong-labeled data (LIDC\_PM, N=72). We repeated the analysis 100 times to measure the statistical variance of the models. Table~\ref{table:spiculation_results_seg} shows the classification results of each model, and their external validation. The 10x10-fold CV of the model using Size and our spiculation features (Size+Spiculations) on LIDC\_PM (accuracy=82.6\% and AUC=0.85), which did not use weak-labeled data, outperformed the previous model (Size+Texture, accuracy=74.9\% and AUC=0.83). However, their external validation on LUNGx was not good as the CV results (Size+Texture: accuracy=66.4\% and AUC=0.63, Size+Spiculation: accuracy=67.5\% and AUC=0.65). The Size+Spiculations model trained by weak-labeled data showed comparable performance (accuracy=75.2\% and AUC=0.80) to the Size+Texture model (accuracy=73.7\% and AUC=0.82) in the validation on LIDC\_PM, but the performance of Size+Spiculations was much higher (accuracy=71.8\% and AUC=0.76) than Size+Texture (accuracy=57.8\% and AUC=0.61) in the external validation.


\begin{table*}[h]
\centering
\caption{Malignancy classification results. Size: BB\_AP, Texture: SD\_IDM, and Spiculations: $N_s$, $N_a$, $N_l$, $N_p$, $r_a$, $s_1$, and $s_2$.}
\label{table:spiculation_results_seg}
\setlength{\tabcolsep}{6pt}
\begin{tabular}{rlcccc}
\hline
&Features & Sensitivity & Specificity & Accuracy & AUC\\
\hline
\multicolumn{6}{l}{\textit{10x10-fold CV on LIDC\_PM}}\\
 & Size+Texture & 77.1$\pm$2.2\% & 71.9$\pm$2.4\% & 74.9$\pm$1.4\% & 0.83$\pm$0.01\\
 & Size+Spiculation & 84.4$\pm$1.0\% & 80.2$\pm$2.7\% & 82.6$\pm$1.1\% & 0.85$\pm$0.01\\
\hline
\multicolumn{6}{l}{\textit{Validation on LIDC\_PM}}\\
&Size+Texture & 73.4$\pm$0.7\% & 74.2$\pm$0.1\% & 73.7$\pm$0.4\% & 0.82$\pm$0.01\\
&Size+Spiculations & 76.0$\pm$0.9\% & 74.2$\pm$0.1\% & 75.2$\pm$0.5\% & 0.80$\pm$0.01\\
\hline
\multicolumn{6}{l}{\textit{External validation on LUNGx}}\\
&Size+Texture & 75.3$\pm$4.5\% & 40.8$\pm$4.2\% & 57.8$\pm$2.7\% & 0.61$\pm$0.03\\
&Size+Spiculations & 77.6$\pm$5.8\% & 66.2$\pm$6.7\% & 71.8$\pm$2.4\% & 0.76$\pm$0.01\\
\hline
\end{tabular}
\end{table*}

\begin{table*}[h]
\small
\centering
\caption{Comparison with the top 3 participants and 6 radiologists in LUNGx Challenge \citep{armato2016lungx}}
\label{table:lungx_summary}
\begin{tabular}{rrm{1.8cm}m{5cm}m{1.8cm}}
\hline
 & AUC & Segmentation & Classifier & Training data \\
\hline
\multicolumn{5}{l}{\textit{Models}}\\
9 & 0.61 & Auto & SVM & NLST\\ 
10 & 0.66 &  & Discriminant function & LUNGx\\
Choi et al.~\cite{choi2018medphy} & 0.67 & Manual & SVM - Size+Texture & LIDC\_PM \\
11 & 0.68 & Semi-auto & Support vector regressor & In-house\\
Choi et al.~\cite{choi2018medphy} & 0.68 & Manual & SVM - Size+Texture & LIDC\_RM \\
Proposed Method& 0.69 & Semi-auto & SVM - Size+Spiculation & LIDC\_PM \\
\hline
\multicolumn{5}{l}{\textit{Radiologists}}\\
 1 & 0.70 \\
 2 & 0.75 \\
\textbf{Proposed Method} & \textbf{0.76} & \textbf{Semi-auto} & \textbf{SVM - Size+Spiculations} & \textbf{LIDC\_RM} \\
 3 & 0.78 \\
 4 & 0.82 \\
 5 & 0.83 \\
 6 & 0.85 \\
\hline
\end{tabular}
\end{table*}

Table~\ref{table:lungx_summary} shows the comparisons with the top 3 participants and 6 radiologists in LUNGx Challenge \citep{armato2016lungx}. The model trained using the weak-labeled data showed an AUC of 0.76, which was better than the best model and two radiologists in the LUNGx challenge. The model trained by the strong-labeled data (AUC=0.69). Our previous radiomics model (Size+Texture) showed comparable performance (strong-labeled: AUC=0.67 and weak-labeled: AUC=0.68) with the best model in the LUNGx Challenge (AUC=0.68). Weak-labeled data training generated more robust and flexible models due to the larger volume of data available (about ten times larger than the strong-labeled data); even though the outcomes were not pathologically proven, they were correlated with the real outcome.

\section{Discussion} \label{sec:discussions}
Our reproducible and interpretable number-of-spiculations feature ($N_s$) achieved higher correlation with RS than other features, as shown in Table \ref{table:good_features_seg}. Many texture features were selected in the top 20, but since it is hard to interpret the correlation between these texture features and spiculations, these can be excluded to avoid the uninterpretability of the final models. Moreover, just using our interpretable features, we can also avoid the mandatory image/feature harmonization in the pre-processing step for any new given dataset (repository). Roundness features showed good correlations with spiculation as well as good accuracy in the malignancy prediction because they quantify the irregularity of the target shape. However, these cannot filter out lobulation or attachments from spiculations.

The radiomics models using semi-auto segmentation showed relatively lower performance than manual segmentation. The models using size (BB\_AP) and texture (SD\_IDM) showed a big difference between manual segmentation (79.2\% accuracy) and semi-auto segmentation (73.7\% accuracy). However, it is difficult to normalize the texture feature. Thus the models using SD\_IDM were less stable, and the performance was significantly degraded in the weak-labeled data training and external validation.

Adding radiologist's spiculation score into our previous radiomics model using size and texture (Size+Texture, 74.9\% accuracy) \citep{choi2018medphy} could improve the performance (Size+Texture+RS, 77.4\% accuracy). Similarly, combining Size and RS without Texture (Size+RS, 76.5\% accuracy) showed better performance, and A model combining our spiculation features and Size without texture (Size+Spiculations, 75.2\% accuracy) was slightly better than Size+Texture. In essence, the texture feature SD\_IDM could be replaced by our interpretable spiculation features.

We employed weak-labeled data (LIDC\_RM) to train the malignancy prediction model because of the lack of pathological malignancy data. These models showed comparable performance to the model trained by strong-labeled data (LIDC\_PM).
In the case of strong-labeled data training, it was difficult to avoid bias and over-fitting due to the lack of training data, while building accurate prediction models for malignant nodules. Hence, these models were more susceptible to failure because of the lack of adaptability to out-of-training unlabeled data.
In contrast, weak-labeled data training can help build models that mimic conventional lung cancer screening by radiologists in the clinic using correlation with pathological malignancy. Moreover, a large amount of weak-labeled training data is usually accessible, thus allowing the creation of a more robust model and better performance than the strong-labeled data in external validation.

Therefore, we provide guidelines for a radiomics workflow to overcome the limitations of conventional radiomics studies using weak-labeled data and interpretable and reproducible features. Specifically, if the number of strong-labeled datasets is insufficient to build a good model, the abandoned weak-labeled data can be utilized in further analysis, as in the current study. Leveraging weak-labeled data in the clinic enables continuous tuning of radiomics models -- training using diagnosis (weak-labeled) followed by evaluation using the clinical outcomes (strong-labeled). A possible pipeline for the new radiomics is as follows:
\begin{enumerate}
    \item Univariate analysis or unsupervised learning of strong-labeled data
    \item Build multivariate models based on results from step 1 and cross-validation using the data
    \item Univariate analysis or unsupervised learning of weak-labeled data
    \item Enhance the model from step 2 based on the results from step 3
    \item External validation
    \item Repeat steps 3--5 to tune the model
\end{enumerate}


\subsection{Conclusion and Future Work} \label{sec:conclusion}
We developed a reproducible and interpretable, parameter-free technique for quantifying spiculations on nodules using the area distortion metric from the conformal (angle-preserving) spherical parameterization. In this paper, to the best of our knowledge for the first time, we exploit the insight that for an angle-preserved (conformal) spherical mapping of a given nodule, the negative area distortion precisely characterizes the spiculations/spikes on that nodule. The spiculation quantification measures and radiomics features based on reproducible semi-automatic segmentation of nodule was then applied to the radiomics framework for pathological malignancy prediction. The number-of-spiculations feature was found to be highly correlated (Spearman's rank correlation coefficient $\rho = 0.44$) with the radiologists' spiculation score. Using just our interpretable features (size, attachment, spiculation, lobulation) in the radiomics framework, we were able to achieve AUC$=$0.80 on LIDC and AUC$=$0.76 on LUNGx (the previous LUNGx best being AUC$=$0.68).

In the future, we will exhaustively test our reproducible and interpretable model for lung cancer screening in the clinic. We plan to apply the recently developed deep learning models to segment nodules and expect further improvement in the spiculation quantification as well as prediction accuracy. The proposed interpretable features will be further investigated in other cancers, e.g., breast cancer BI-RADS \citep{kerlikowske2018automated}.


\section*{Acknowledgements}
This project was supported in part by NIH/NCI Grant R01CA172638, AFOSR Grants FA9550-17-1-0435, and FA9550-20-1-0029, National Institute of Aging Grant R01-AG048769, MSK Cancer Center Support Grant/Core Grant (P30 CA008748), and a grant from Breast Cancer Research Foundation BCRF-17-193. None of the authors have any competing financial interests. All the datasets used in this study are publicly available.

\newpage
\onecolumn
\appendix

\section{Results using manual segmentation}
\label{sec:appendix}
We also evaluated our method using manual segmentation. The consensus manual contour generated by using STAPLE to combine multiple contours (up to 4). Table~\ref{table:good_features_man} shows the univariate analysis results of the top 20 features using manual segmentation. None of Dhara's spiculation scores ($s_a$ and $s_b$) were selected in the top 20 features even though they were significant features. Table~\ref{table:spiculation_results_man} shows the malignancy classification results of each model, and their external validation. The 10x10-fold CV of the model using Size and our spiculation features (Size+Spiculations) on LIDC\_PM  (accuracy=85.1\% and AUC=0.85), which did not use weak-labeled data, showed comparable performance to the previous model (Size+Texture, accuracy=84.9\% and AUC=0.89). The Size+Spiculations model trained by weak-labeled data showed comparable performance (accuracy=79.3\% and AUC=0.86) to the Size+Texture model (accuracy=79.2\% and AUC=0.83) in the validation on LIDC\_PM, but the performance of Size+Spiculations was much higher (accuracy=71.5\% and AUC=0.74) than Size+Texture (accuracy=60.7\% and AUC=0.68) in the external validation.

\begin{table*}[h!]
\centering
\caption{Twenty highly correlated features with RS in univariate analysis using manual segmentation. $T_s$: the threshold to binaries the RS}
\vspace{3mm}
\label{table:good_features_man}
\setlength{\tabcolsep}{2pt}
\begin{tabular}{rlccccr}
\hline
Rank & Feature name & \multicolumn{4}{c}{AUC} & Corr\\
 &  & $T_s>1$ & $T_s>2$ & $T_s>3$ & Average & $\rho$\\
\hline
1 & Roundness(mesh) & 0.73 & 0.84 & 0.82 & 0.80 & -0.44\\
2 & SD LRE & 0.74 & 0.80 & 0.78 & 0.77 & 0.44\\
3 & $N_s$ & 0.72 & 0.82 & 0.83 & 0.79 & 0.44\\
4 & Mean $\epsilon$ & 0.71 & 0.80 & 0.79 & 0.77 & -0.40\\
5 & Minimum $\epsilon$ & 0.71 & 0.80 & 0.77 & 0.76 & -0.40\\
6 & WPM1 & 0.72 & 0.75 & 0.77 & 0.74 & 0.40\\
7 & Median $\epsilon$ & 0.71 & 0.78 & 0.79 & 0.76 & -0.39\\
8 & 2D Roundness(voxel) & 0.69 & 0.82 & 0.79 & 0.77 & -0.38\\
9 & 2D WPM1 & 0.70 & 0.74 & 0.75 & 0.73 & 0.37\\
10 & SD LRLGE & 0.69 & 0.74 & 0.74 & 0.72 & -0.35\\
11 & Mean LRLGE & 0.69 & 0.74 & 0.72 & 0.72 & -0.35\\
12 & 2D Sum & 0.69 & 0.73 & 0.72 & 0.71 & -0.34\\
13 & SD CS & 0.69 & 0.70 & 0.68 & 0.69 & -0.34\\
14 & $s_1$ & 0.67 & 0.77 & 0.75 & 0.73 & -0.33\\
15 & SD LGRE & 0.67 & 0.71 & 0.71 & 0.70 & -0.31\\
16 & SD SRLGE & 0.66 & 0.70 & 0.70 & 0.69 & -0.30\\
17 & $N_l$ & 0.65 & 0.69 & 0.71 & 0.68 & 0.29\\
18 & Mean Energy & 0.67 & 0.64 & 0.63 & 0.65 & -0.29\\
19 & $s_2$ & 0.65 & 0.74 & 0.72 & 0.70 & 0.28\\
20 & SD SRE & 0.64 & 0.71 & 0.70 & 0.69 & -0.27\\
\hline
\end{tabular}
\end{table*}


\begin{table*}[h!]
\centering
\caption{Malignancy classification results. Size: BB\_AP, Texture: SD\_IDM, and Spiculations: $N_s$, $N_a$, $N_l$, $N_p$, $r_a$, $s_1$, and $s_2$.}
\label{table:spiculation_results_man}
\setlength{\tabcolsep}{6pt}
\begin{tabular}{rlcccc}
\hline
&Features & Sensitivity & Specificity & Accuracy & AUC\\
\hline
\multicolumn{6}{l}{\textit{10x10-fold CV on LIDC\_PM}}\\
 & Size+Texture & 86.9$\pm$1.0\% & 81.2$\pm$3.6\% & 84.4$\pm$1.7\% & 0.89$\pm$0.01\\
 & Size+Spiculation & 87.9$\pm$1.0\% & 81.5$\pm$1.6\% & 85.1$\pm$1.0\% & 0.88$\pm$0.01\\
\hline
\multicolumn{6}{l}{\textit{Validation on LIDC\_PM}}\\
&Size+Texture & 78.1$\pm$0.3\% & 80.7$\pm$0.3\% & 79.2$\pm$0.2\% & 0.86$\pm$0.01 \\
&Size+Spiculations & 78.3$\pm$0.7\% & 80.6$\pm$0.0\% & 79.3$\pm$0.4\% & 0.83$\pm$0.01\\
\hline
\multicolumn{6}{l}{\textit{External validation on LUNGx}}\\
&Size+Texture & 67.5$\pm$5.7\% & 54.0$\pm$5.5\% & 60.7$\pm$3.5\% & 0.68$\pm$0.04\\
&Size+Spiculations & 81.9$\pm$1.9\% & 61.3$\pm$3.7\% & 71.5$\pm$2.1\% & 0.74$\pm$0.02\\
\hline
\end{tabular}
\end{table*}

\end{document}